\documentclass{ieeetj}
\usepackage{cite}
\usepackage{amsmath,amssymb,amsfonts}
\usepackage{algorithmic}
\usepackage{graphicx,color}
\usepackage{textcomp}
\usepackage{xcolor}
\usepackage{hyperref}
\hypersetup{hidelinks=true}
\usepackage{algorithm,algorithmic}

\usepackage{setspace}
\usepackage{float} 
\usepackage{bm}
\usepackage{url}
\usepackage{acronym} 
\usepackage{array}
\usepackage[caption=false,font=normalsize,labelfont=sf,textfont=sf]{subfig}
\usepackage{stfloats}
\usepackage{verbatim}

\def\BibTeX{{\rm B\kern-.05em{\sc i\kern-.025em b}\kern-.08em
    T\kern-.1667em\lower.7ex\hbox{E}\kern-.125emX}}
\AtBeginDocument{\definecolor{tmlcncolor}{cmyk}{0.93,0.59,0.15,0.02}\definecolor{NavyBlue}{RGB}{0,86,125}}

\def\OJlogo{\vspace{-4pt}$<$Society logo(s) and publication title will appear here.$>$}
\def\seclogo{\vspace{10pt}$<$Society logo(s) and publication title will appear here.$>$}

\def\authorrefmark#1{\ensuremath{^{\textbf{#1}}}}

\usepackage{xparse}

\NewDocumentCommand\bbm{}{ \begin{bmatrix} }
\NewDocumentCommand\ebm{}{ \end{bmatrix} }

\NewDocumentCommand\Matrix{m}{ \boldsymbol{\mathbf{#1}} }

\NewDocumentCommand\LieGroupSE{m}{ \mathrm{SE}(#1) }

\NewDocumentCommand\ZeroMatrix{}{ \Matrix{0} }
\NewDocumentCommand\IdentityMatrix{}{ \Matrix{1} }

\NewDocumentCommand\CoordinateFrame{m}{ \underrightarrow{\Matrix{\mathcal{F}}}_{#1} }

\newcommand{\uk}{\mathbf{u}_k}
\newcommand{\du}{\mathbf{\delta u}}

\newcommand{\dubar}{\mathbf{\delta \bar{u}}}
\newcommand{\duk}{\mathbf{\delta u}_k}
\newcommand{\ukop}{\mathbf{u}_{k,op}}
\newcommand{\Puk}{(\mathbf{P}^T\mathbf{u}_k)}
\newcommand{\Pukop}{(\mathbf{P}^T\mathbf{u}_{k,op})}
\newcommand{\Pudk}{(\mathbf{P}^T\mathbf{\delta u}_{k})}

\newcommand{\T}{\Matrix{T}}
\newcommand{\Tk}{\Matrix{T}_k}
\newcommand{\Trefk}{\Matrix{T}_{\mathrm{ref},k}}

\newcommand{\dTk}{\delta \Matrix{T}_{k}}
\newcommand{\Tkop}{\Matrix{T}_{k,\mathrm{op}}}
\newcommand{\Tkone}{\Matrix{T}_{k+1}}
\newcommand{\Tkoneop}{\Matrix{T}_{k+1,\mathrm{op}}}

\newcommand{\ek}{\Matrix{\epsilon}_k}
\newcommand{\ekone}{\Matrix{\epsilon}_{k+1}}
\newcommand{\ekop}{\Matrix{\epsilon}_{k,\mathrm{op}}}
\newcommand{\ebar}{\Matrix{\epsilon}}
\newcommand{\ebarop}{\Matrix{\epsilon}_\mathrm{op}}

\newcommand{\ub}{\Matrix{u}}
\newcommand{\uop}{\Matrix{u}_{\mathrm{op}}}
\newcommand{\y}{\Matrix{y}} 
\newcommand{\yk}{\Matrix{y}_{k}}
\newcommand{\yop}{\Matrix{y}_{\mathrm{op}}}
\newcommand{\ykop}{\Matrix{y}_{k,\mathrm{op}}}
\newcommand{\dy}{\delta \Matrix{y}}
\newcommand{\dyk}{\delta \Matrix{y}_{k}}

\newcommand{\db}{\Matrix{d}}

\newcommand{\vel}{\Matrix{\Phi}_{\mathrm{vel}}(\ub)}
\newcommand{\velop}{\Matrix{\Phi}_{\mathrm{vel}}(\uop)}
\newcommand{\lat}{\Matrix{\Phi}_{\mathrm{lat}}(\y)}
\newcommand{\latop}{\Matrix{\Phi}_{\mathrm{lat}}(\yop)}

\newcommand{\V}{\Matrix{V}}
\newcommand{\W}{\Matrix{W}}
\newcommand{\R}{\Matrix{R}}
\newcommand{\Q}{\Matrix{Q}}
\newcommand{\Hb}{\Matrix{H}}
\newcommand{\F}{\Matrix{F}}
\newcommand{\G}{\Matrix{G}}
\newcommand{\M}{\Matrix{M}}
\newcommand{\N}{\Matrix{N}}
\newcommand{\A}{\Matrix{A}}
\newcommand{\B}{\Matrix{B}}

\newcommand{\specialcell}[2][c]{%
	\begin{tabular}[#1]{@{}c@{}}#2\end{tabular}}

\acrodef{BIT*}{Batch Informed Trees}
\acrodef{DOF}{Degrees of Freedom}
\acrodef{DRDC}{Defence Research and Development Canada}
\acrodef{DWA}{Dynamic Window Approach}
\acrodef{GPS}{Global Positioning System}
\acrodef{GMM}{Gaussian Mixture Model}
\acrodef{GUI}{Graphical User Interface}
\acrodef{LiDAR}{Light Detection and Ranging}
\acrodef{LPA*}{Lifelong Planning A*}
\acrodef{MPC}{Model Predictive Control}
\acrodef{OGM}{Occupancy Grid Maps}
\acrodef{PID}{Proportional-Integral-Derivative}
\acrodef{RHC}{Receding Horizon Control}
\acrodef{RMS}{Root Mean Squared}
\acrodef{ROC}{Radius of Curvature}
\acrodef{ROS2}{Robot Operating System 2}
\acrodef{RRT}{Rapidly Exploring Random Trees}
\acrodef{STEAM}{Simultaneous Trajectory Estimation and Mapping}
\acrodef{TEB}{Timed Elastic Band}
\acrodef{UGV}{Unmanned Ground Vehicle}
\acrodef{VTR}[VT\&R]{Visual Teach \& Repeat}
\acrodef{VTR3}[VT\&R3]{Visual Teach \& Repeat 3}

\begin{document}
\receiveddate{XX Month, XXXX}
\reviseddate{XX Month, XXXX}
\accepteddate{XX Month, XXXX}
\publisheddate{XX Month, XXXX}
\currentdate{XX Month, XXXX}
\doiinfo{XXXX.2022.1234567}

\markboth{}{Author {et al.}}

\title{Off the Beaten Track: Laterally Weighted Motion Planning for Local Obstacle Avoidance}

\author{Jordy Sehn\authorrefmark{1}, Timothy D. Barfoot\authorrefmark{1}, Fellow, IEEE,\\ and Jack Collier \authorrefmark{2}}
\affil{Institute for Aerospace Studies, University of Toronto, Toronto, ON Canada}
\affil{Defence Research and Development Canada, Suffield, AB, Canada}
\corresp{Corresponding author: Jordy Sehn (email: jordy.sehn@robotics.utias.utoronto.ca).}
\authornote{This research is supported by the Natural Sciences and
	Engineering Research Council of Canada (NSERC) and the
	Vector Scholarship in Artificial Intelligence.}

\begin{abstract}
We extend the behaviour of generic sample-based motion planners to support obstacle avoidance during long-range path following by introducing a new edge-cost metric paired with a curvilinear planning space. The resulting planner generates naturally smooth paths that avoid local obstacles while minimizing lateral path deviation to best exploit prior terrain knowledge from the reference path. In this adaptation, we explore the nuances of planning in the curvilinear configuration space and describe a mechanism for natural singularity handling to improve generality. We then shift our focus to the trajectory-generation problem, proposing a novel \ac{MPC} architecture to best exploit our path planner for improved obstacle avoidance. Through rigorous field robotics trials over 5 km, we compare our approach to the more common direct path-tracking \ac{MPC} method and discuss the promise of these techniques for reliable long-term autonomous operations.
\end{abstract}



\maketitle

\section{INTRODUCTION}

Robot navigation in unstructured outdoor environments is a challenging-yet-critical task for many mobile robotics applications including transportation, mining, and forestry. In particular, robust localization in the presence of both short- and long-term scene variations without reliance on a \ac{GPS} becomes particularly difficult. Furthermore, the off-road terrain-assessment problem is non-trivial to generalize as the variety of potential obstacles increases, all of which require careful identification, planning, and control to prevent collisions.

\ac{VTR} \cite{Furgale2010} tackles these problems by suggesting that often it is sufficient for a mobile robot to operate on a network of paths previously taught by a human operator. During a learning phase (the \textit{teach pass}) the robot is manually piloted along a route whilst building a visual map of the environment using a rich sensor such as a stereo camera. In the autonomous traversal phase (the \textit{repeat pass}), live stereo images are used to localize to the map with high precision and resiliency to lighting and seasonal changes \cite{Paton2016,Gridseth2021}.

\begin{figure*}[!t]
	\centering
	\includegraphics[scale=0.68]{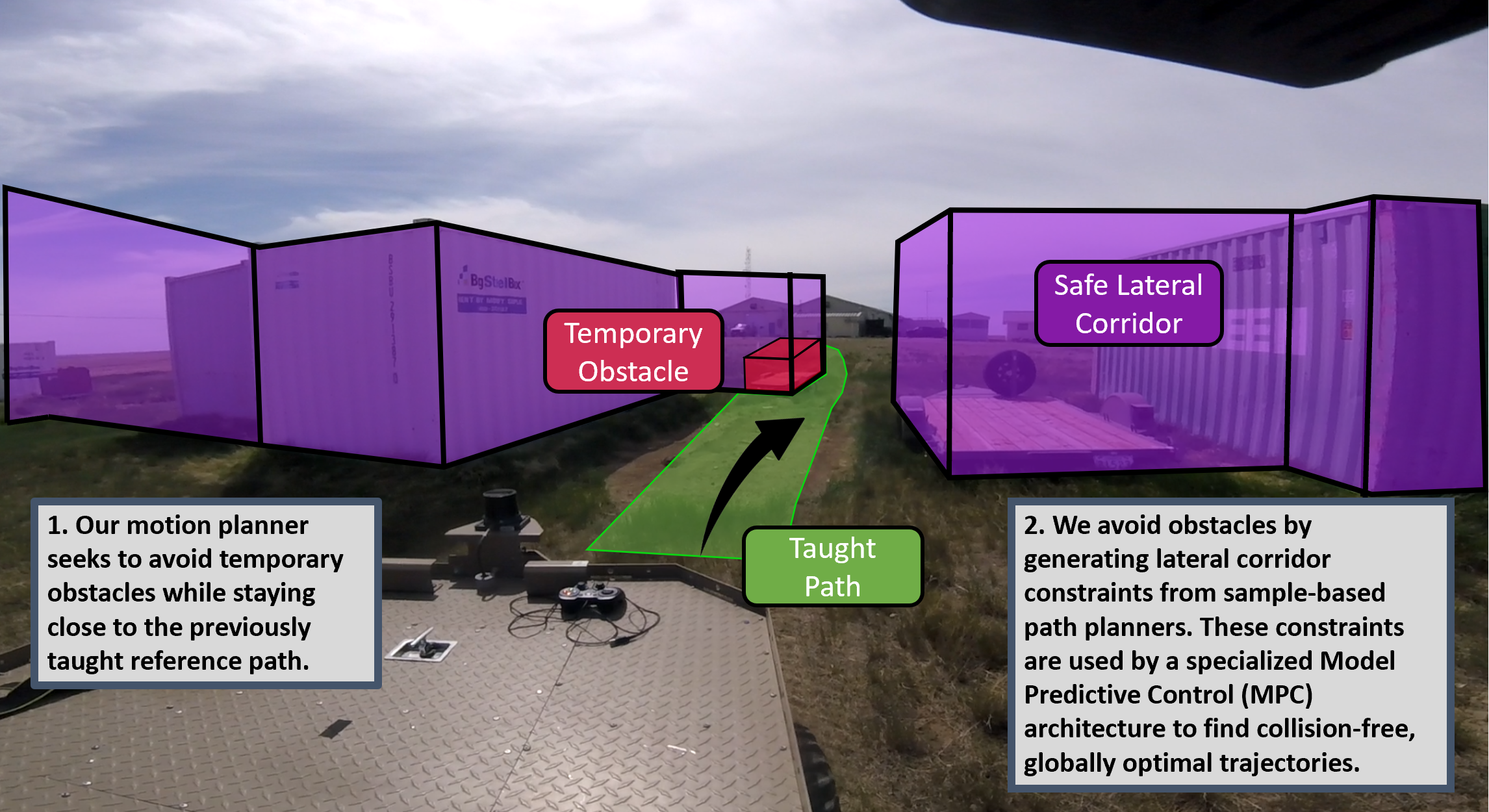}
	\caption[Motion Planner Overview]{We present an architecture for locally avoiding unmapped obstacles along a reference path by extending sample-based motion planners to encourage trajectories with characteristics that best exploit prior path knowledge. The goal of our planner is to follow the previously established reference path as closely as possible while avoiding any new obstacles that may have appeared along the path. To accomplish this task we use sample-based motion planners deployed in a laterally constrained corridor configuration space to find candidate collision-free paths, and then apply Model Predictive Control (MPC) to generate optimized trajectories for the robot. Our system is validated on an ARGO Atlas J8 robot in real-world scenarios.}
	\label{Motion_Planner_Overview}
	\vspace{-1em}
\end{figure*}

In practice, this architecture works well to address both the localization and terrain-assessment problems without using GNSS/GPS. By having a human operator drive the teach pass, we exploit the strong prior that the original path was traversable. It follows that, at least in the short term, it is likely the taught path remains collision free and by following the path closely in the repeat, minimal terrain assessment is required.

With the advent of \ac{LiDAR} implementations of teach and repeat \cite{McManus2012,Krusi2014,Cherubini2013} whose localization system is less sensitive to viewpoint changes when repeating a path than stereo, we explore the possibility of temporarily deviating from the teach path to avoid new obstacles and increase the practicality of VT\&R. In this application, we highlight the reliability of change detection for obstacle perception in difficult environments. Additionally, we emphasize the importance of path planning that minimizes deviations from the taught path to take advantage of the human terrain assessment prior whenever possible. This allows us to navigate in the safest possible manner in the event of terrain-assessment errors.


The primary contribution of our system is in the applied planning domain where we present a novel edge-cost metric that, when combined with a curvilinear planning space, extends the capabilities of a generic sample-based optimal motion planner to generate paths that naturally avoid local obstacles along the teach pass.  Notably, in obstacle-free environments, our planner ensures that the solution remains on the original teach path without cutting corners. The key feature of our edge-cost metric is its ability to encourage path solutions that strike a delicate balance between minimizing path length and lateral path deviation, all achieved without relying on waypoints. Additionally, we demonstrate that our new metric maintains the fundamental properties of the underlying planning algorithm by utilizing an admissible sampling heuristic. 


After generating suitable obstacle-avoidance plans, we propose two distinct control paradigms for trajectory generation using \ac{MPC}. In the first case, we directly track the planner output paths while utilizing \ac{MPC} to enforce basic kinematic and acceleration constraints. This approach generates smooth trajectories that effectively avoid local obstacles. In the second implementation, we decouple the \ac{MPC} from the planner by explicitly following the taught reference path instead of the planner output. To avoid obstacles, the homotopy class of the current path solution is used to enforce a set of corridor state constraints, ensuring a collision-free and robust \ac{MPC} trajectory.




In this work, we expand and improve upon the motion-planning methodology and results of our previous conference publication \cite{Sehn2022}. The key contributions of our extension are as follows:

\begin{enumerate}
	\item \textbf{Rotation Singularity Handling}: We introduce a mechanism for managing rotation singularities within the path planner, enhancing the overall completeness of the planning framework.
	
	\item \textbf{Homotopy-Class-Guided \ac{MPC}}: We propose and evaluate a novel homotopy-class-guided \ac{MPC} that leverages the strengths of our path planner, providing a significant advancement over our previous direct-tracking \ac{MPC} implementation.
	
	\item \textbf{Experimental Validation}: We conduct a series of autonomous navigation experiments in unstructured, GPS-denied environments using a new robotic platform, demonstrating the efficacy of our approach.
\end{enumerate}

To illustrate the effectiveness of our path planner, we conducted extensive simulations across a diverse range of realistic path-following scenarios. Notably, in 87\% of our randomly generated trials, our planner successfully identified a path to the goal, maintaining a maximum lateral deviation of 2.5 meters from the original reference path. In all instances of path failure, our planner accurately reported that no solution could be found that met the path deviation constraints due to the random arrangement of obstacles in the environment.

Additionally, we performed long-term autonomous navigation experiments covering over 5 kilometers in various unstructured environments, demonstrating the system's robust obstacle-avoidance capabilities and its potential for real-world deployment. Our homotopy-guided \ac{MPC} motion planner successfully navigated a large ARGO Atlas J8 robot through increasingly complex obstacle avoidance scenarios, achieving a 100\% obstacle avoidance rate with no operator interventions, while minimizing lateral path deviation compared to direct-tracking \ac{MPC} approaches.

The remainder of the paper is organized as follows: In Section II, we provide a comprehensive overview of the literature on path following with local trajectory planning. Section III introduces our sample-based motion planner based on \ac{BIT*} \cite{Gammell2015}, highlighting our modifications for planning in curvilinear configuration spaces. Section IV addresses the handling of corner-case singularities that may arise from this approach. Following the path planner, Section V presents our homotopy-class-guided \ac{MPC} for controlling the autonomous vehicle. We then provide a series of comprehensive simulation experiments and results in Section VI, followed by our field trial experiments in Section VII. Finally, we analyze and discuss the key findings of our results in Section VIII.

\section{Related Work}
\label{related_work}

\subsection{Path Planning in Teach and Repeat}
Path planning for local obstacle avoidance is a rich and well studied field with a wide range of approaches. In this work, we are particularly concerned with local obstacle avoidance for unmanned ground vehicles with fixed underlying reference plans, and as such limit our review of the literature to this subsection of obstacle planning. Given the current state of the robot and known positions of obstacles, we explore the methods for generating trajectories that avoid the obstacle and return to the reference plan.


The emergence of 3D \ac{LiDAR}-based \ac{VTR} architectures \cite{McManus2012,Baril2021,Wu2022,Krusi2014} with improved localization robustness up to several meters away from the path has made local obstacle avoidance more feasible. Krusi et al. \cite{Krusi2014} utilized the additional mobility freedom to navigate reliably in dynamic environments. They actively detected potential obstacles by analyzing irregularities in the distance between concentric \ac{LiDAR} rings, assuming local planarity of the ground. Once detected, an online motion planner generated a tree of many potential trajectories around the obstacles, selecting the one that best satisfied system constraints. This approach improved the long-term autonomy of the teach-and-repeat architecture but was limited to mostly structured environments with simple obstacle geometries. In 2016, the same team attempted to improve upon this approach by incorporating a traditional local planner based on \ac{RRT} \cite{Krusi2016} with dynamic waypoint generation. However, for path-following applications, the selection of waypoints can be cumbersome.

When working with local obstacle-avoidance planners, a key challenge is determining how much deviation from the global plan should be allowed, or in the case of teach and repeat, how far the robot should be permitted to deviate from the previously taught reference path. Unstructured environments are often filled with unexpected obstacles that may require significant deviation from the original path to find a collision-free solution. Earlier versions of \ac{VTR} \cite{Berczi2016} lacked the ability to plan collision-free trajectories around detected obstacles on the path, instead opting for the conservative approach of stopping and signaling for operator intervention. This method has proven effective in scenarios where a single operator oversees multiple robots, as it reduces the risk of unsupervised operation. In contrast, Krusi et al. \cite{Krusi2016} adopted a more conventional local and global planner schema, where the global planner would replan the reference path more broadly if the local planner failed to find a solution. While this approach increases autonomy, it forgoes the advantages associated with leveraging prior terrain knowledge.

In our application, we take a hybrid approach. Initially, we attempt to find collision-free paths within a localized area, constrained by a maximum lateral boundary around the reference path, chosen based on our localization capabilities. If obstacle avoidance requires exceeding this lateral boundary, no solution is returned, and operator intervention is signaled. We argue that this scenario is unlikely as the odds of encountering extremely large unforeseen obstacles are low due to the strength of the prior terrain knowledge. Nonetheless, this approach prioritizes safety and reliability over a more autonomous, adaptive solution. However, our planner could easily be modified to support larger lateral boundaries in systems that can afford greater path deviations or are not as tightly bound to a fixed reference path in the event of global replanning.

Mattamala et al. \cite{Mattamala2022} explore the use of a locally reactive controller for completing visual teach-and-repeat missions in the presence of obstacles. Their research employs local elevation maps to compute vector representations of the environment and directly generate control twist commands using a Riemannian Motion Policies controller. The approach demonstrates effectiveness in handling numerous small obstacles, benefiting from its efficient computational processing. However, it has shown a tendency to encounter challenges when dealing with more complex obstacle formations, as it relies on maneuvering through Signed Distance Fields and can fall into local minima.


More recently, there has been considerable effort in exploring alternative navigation strategies beyond classical approaches. For example, Meng et al. \cite{Meng2021} demonstrate the use of a deep network to learn high-level navigation behaviours from a dataset of example demonstrations. Their network takes in a sequence of images from a `teach' dataset and outputs a short-distance waypoint ahead of the robot based on the current observation. While not explicitly designed for obstacle avoidance, this approach naturally exhibits avoidance behaviour in addition to path following. Although promising, the path-following errors in these techniques \cite{Kumar2021,Dugas2022,Sadek2022} are currently larger than what is acceptable for precise teach-and-repeat applications.

None of the aforementioned works explicitly preserve consistent viewpoints or limit lateral path deviations during obstacle avoidance. These factors are crucial for mitigating localization failures during repeats.

\subsection{Local Trajectory Planning}
Another approach to path following with local obstacle avoidance is presented by Liniger et al. \cite{Liniger2015}. They propose a decoupled path planner and \ac{MPC} architecture, applied in the domain of autonomous racing. 
Their method discretizes the track into a grid of cells and employs dynamic programming \cite{Bellman1952} to identify the shortest path around competing vehicles, leading back to the optimal racing line. From this initial solution, they define lateral corridor constraints that guide an \ac{MPC} controller to find an optimal trajectory within the corridor. As they can define this corridor as convex, this ensures that the \ac{MPC} does not get stuck in local minima and can maintain high control rates. While this approach is powerful and serves as a significant source of motivation for our method, it also has its limitations, particularly in the scalability of the planning approach. The brute-force dynamic programming planner is restricted to structured scenes with a few obstacles of known fixed size.

The guided \ac{MPC} method of motion planning used by Liniger et al. \cite{Liniger2015} is based on the work of Park et al. \cite{Park2015} and has roots in the concept of path homotopies. A path-homotopy class is a mathematical concept used in topology to classify paths based on their topological equivalence \cite{Subhrajit2017}. Two paths belong to the same homotopy class if they can be transformed into each other by smoothly deforming them while keeping their endpoints fixed. The notion of path homotopy is based on the idea that the shape or topology of a path is more important than its precise geometric details.

The popular \ac{TEB} approach \cite{Rosman2016} similarly leverages path-homotopy classes to find robust planning solutions. In their implementation, \ac{TEB} utilizes Voronoi diagrams to span the environment and discover as many homotopy classes as possible. They then use analysis based on the related concept of path homology to refine and filter these sets of paths, followed by parallel optimization to select the optimal solution. This approach offers improved scalability for real-time applications. However, it does not inherently support smooth path following without relying on waypoints and is computationally expensive due to the exhaustive nature of Voronoi diagram generation. Bhattacharya and Likhachev \cite{Bhattacharya2010} perform a similar exhaustive homotopy class search, instead using the graph-based search algorithm A* \cite{Hart1968} to improve the scalability.


Our approach to planning within a curvilinear configuration space is similar to motion planners that utilize the Fren\'et Frame \cite{Manfredo1976}, which is commonly employed in self-driving cars to plan paths relative to road centerlines. The Fren\'et Frame is a local, moving coordinate system that follows a reference path, typically defined by a road centerline. It consists of three orthogonal vectors: the tangent vector (pointing along the direction of the path), the normal vector (indicating the direction of curvature), and the binormal vector (perpendicular to the plane of the curve). This coordinate system allows the planner to decompose vehicle motion into longitudinal and lateral components, simplifying path following and obstacle avoidance by reducing the complexity of planning in global coordinates.

Werling et al. (2010) demonstrated how traffic-avoiding trajectories can be sampled in the Fren\'et Frame and jointly optimized for both lateral and longitudinal cost functions, enabling self-driving vehicles to execute various maneuvers that mimic human driving behavior, even at high speeds. Their method leverages vehicle dynamics, including maximum turning radii, to reduce the search space of feasible trajectories. However, the authors acknowledge that this approach faces challenges during extreme maneuvers, particularly when excessive path curvature creates singularity regions in the Fren\'et Frame. While this issue can be safely ignored in most self-driving car applications, where smooth centerline paths and constrained vehicle dynamics dominate, the same cannot be said for smaller differential-drive robots operating in off-road, unstructured environments over more complex reference paths.

In these environments, it may be desirable to take advantage of tighter turns in more complex terrain. Our approach, while not immune to the singularity problem, introduces a novel mechanism to plan effectively within configuration spaces that may contain singularity regions. We achieve this through a preprocessing step that allows us to handle these problematic areas more elegantly, ensuring robust path planning in a wider range of conditions.

Unlike the previous architectures, we elect to use a sample-based planner to identify a strong candidate path that avoids obstacles and combine this result with \ac{MPC} in two ways. The first is a direct tracking \ac{MPC} approach where the output of the planner is used as a reference trajectory for the cost function. The second method is to implement a homotopy-guided \ac{MPC} with corridor constraints defined by the current planner solution's homotopy class. Using sample-based planners such as Rapidly Exploring Random Trees (RRT*) \cite{Sertac2011}, Fast Marching Trees (FMT) \cite{Janson2013}, Probabilistic Roadmaps (PRM) \cite{Kavraki1996}, or D* \cite{Stentz1994}, potentially allow solutions to be generated more tractably in large unstructured environments. Unlike deterministic search algorithms that exhaustively explore the configuration space, sample-based planners take a probabilistic approach by sampling random points in the space. This sampling strategy allows them to efficiently explore the configuration space without being constrained by its size or complexity \cite{Lajevardy2015} and is better suited for the applications of teach and repeat.

\subsection{Model Predictive Control}

We are not the first to try to combine sample-based planners with MPC. Zhou et al. \cite{Zhou2022} run a variation of the sample-based planner Informed RRT* \cite{Gammell2014} to generate shortest-distance paths avoiding obstacles, and opt to track this path directly with \ac{MPC} under the assumption that it will be collision free. Similar to this work, Al-Moadhen et al. \cite{AlMoadhen2022} use \ac{BIT*} as the sample-based motion planner with the addition of a B-Spline smoothing post-processing step to generate kinematically feasible paths to be tracked with \ac{MPC}.

A fundamental difference between previous works and this research lies in the design of the sample-based path planner. By introducing specific modifications to the planner's fundamental structure, including the incorporation of a specialized configuration space and a novel edge-cost metric that prioritizes minimizing lateral deviation from a reference path, we demonstrate the ability to customize this architecture to best align with the path-following problem structure. 

Our approach utilizes a sample-based planner to address the challenge of solving tractable \ac{MPC} problems in obstacle-dense environments. In our initial approach \cite{Sehn2022}, we prioritize tracking the reference path generated by the planner while enforcing kinematic constraints to ensure a smooth trajectory, similar to other existing methods \cite{Li2022,Xu2021}. This implementation is straightforward and effective in avoiding obstacles when the reference path is collision free. However, a drawback of this method is that if the reference path is not explicitly kinematically feasible, it can lead to small tracking errors that may result in collisions. Unlike the comparable work done by Xu et al. \cite{Xu2021} that requires a post-processing step to smooth plans, our planner naturally generates smooth trajectories, mitigating some of these limitations. Nevertheless, we propose a second alternative \ac{MPC} architecture that addresses this issue by utilizing the planner to generate dynamic lateral corridor state constraints. The \ac{MPC} then uses these constraints to generate trajectories within the corridor homotopy class that guarantee obstacle avoidance as done by Liniger et al. \cite{Liniger2015}. We then compare this approach with the more commonly used direct-tracking method.




\section{Path Planning}
\begin{figure*}[t]
	\centering
	\includegraphics[scale=0.66]{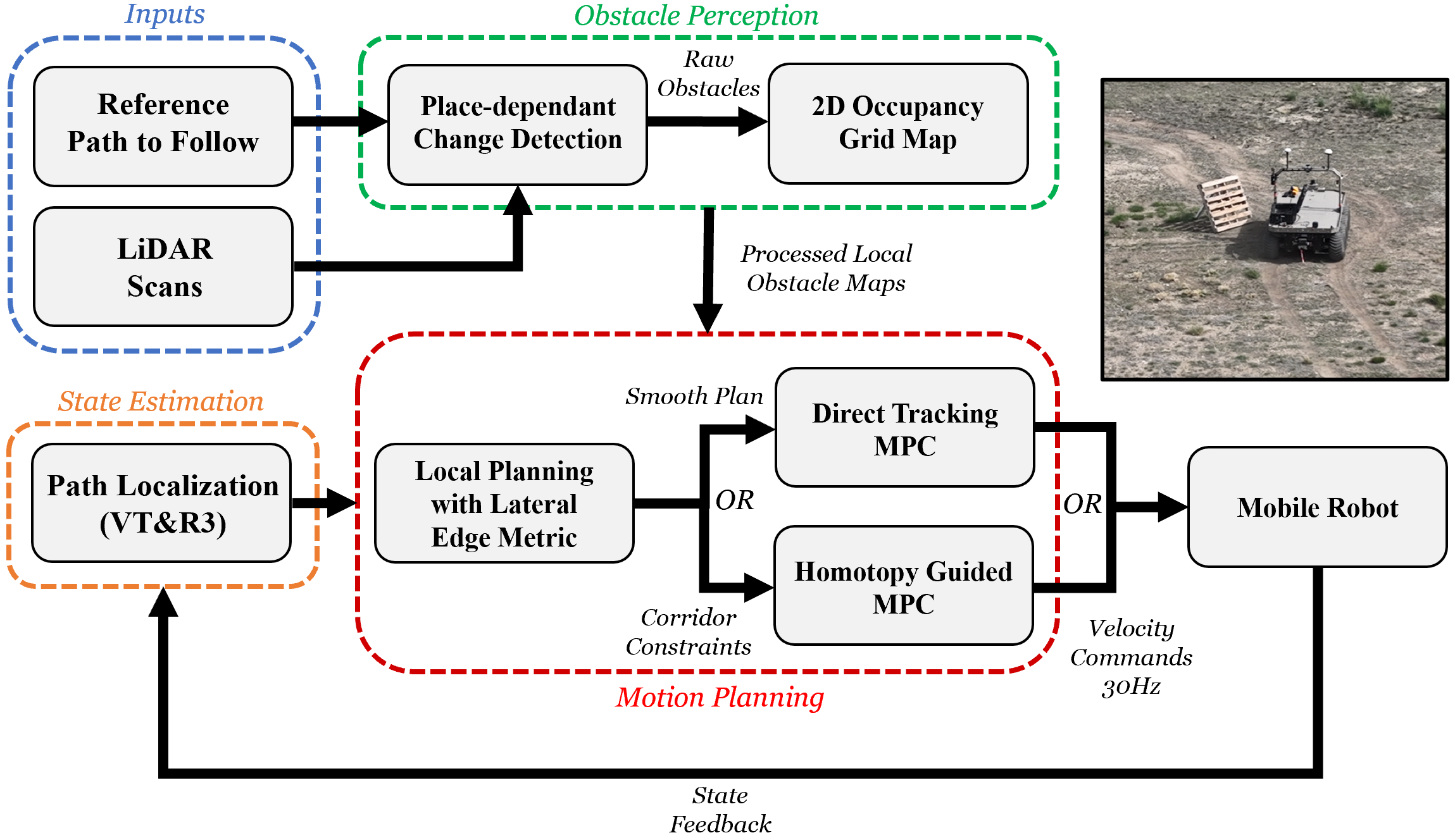}
	\caption{An overview of the proposed obstacle-avoidance system. A change-detection LiDAR perception module identifies previously unmapped structures and updates a locally planar 2D occupancy grid map. The planner finds paths that avoid obstacles using our laterally weighted edge-cost metric and a tracking MPC enforces kinematic constraints on the final trajectory. Velocity commands are then sent to maneuver the robot, at which point we receive new robot state estimates from \ac{VTR} and pass these states to both the planner and MPC to update plans and complete the feedback cycle.}
	\label{figure2}
	\vspace{-1em}
\end{figure*}

This section outlines our approach to autonomously following a previously taught route. To provide context, we present an overview of our complete architecture in Fig. \ref{figure2}, and we will now delve into a detailed description of our local planner. In particular, we draw attention to our two main contributions in this section: the generation of a unique curvilinear coordinate configuration space from an arbitrary reference path and the use of a novel laterally weighted edge cost metric to encourage path solutions that prioritize path following and reducing lateral path deviation in the presence of obstacles.

\subsection{Obstacle Detection}
As the obstacle-detection system is not the focus of evaluation for this work, we direct the reader to Wu \cite{Wu2022} for implementation details. At a high level, we treat obstacle detection as a change-detection problem between the environment's long-term static structures and the current LiDAR observations. If a point from the live scan fails to coincide with a mapped structure, it is likely from a new obstacle, or is representative of a significant change to the environment that should be avoided.

We then adopt classical methods for classifying points in the scan based on a thresholding method similar to Girardeau-Montaut and Marc \cite{Girardeau2005}, while modelling surface roughness with a Gaussian in our classification metric \cite{Berczi2016,Berczi2017}. Going forward we take for granted that all obstacles are reliably detected and projected onto local 2D occupancy grids for collision checking.


In line with typical 2.5D costmap representations of obstacles, we project detected obstacles onto a single plane and discretize the occupancy into a grid of cells. These occupied grid cells are then inflated using the decaying exponential function described by Wu \cite{Wu2022} out to a maximum distance, controlled by a user-adjustable inflation parameter. For our use case, this parameter is set to 30 cm, serving as a buffer around the true position of obstacles to account for sensor measurement errors and unmodeled dynamics during path tracking. The inflation region introduces a trade-off between safety and minimizing lateral deviation from the reference path. In our experiments, we opted for more aggressive settings to demonstrate the planner's capabilities, accepting a higher risk of collisions to highlight its performance potential.




\subsection{Sample-Based Planning Preliminaries and Notation}
While the selection of path planner is arbitrary for the application of our extensions, we employ Batch Informed Trees (BIT*) \cite{Gammell2015} as our baseline planner. BIT* is probabilistically complete, asymptotically optimal, and can be adapted to replan or \textit{rewire} itself in the presence of moving obstacles, making it an ideal candidate for our current and future applications. 

We define our optimal planning problem in the same fashion as Gammell (2015). \textit{Problem Definition 1 (Optimal Planning)}: Let $X \subseteq \mathbb{R}^n$ be the state space of the planning problem, $X_{\text{obs}} \subset X$ be the states in collision with obstacles, and $X_{\text{free}} = X \setminus X_{\text{obs}}$ be the resulting set of permissible states. Let $x_{\text{start}} \in X_{\text{free}}$ be the initial state and $X_{\text{goal}} \subset X_{\text{free}}$ be the set of desired final states. Let $\sigma : [0, 1] \to X$ be a sequence of states (a path) and $\Sigma$ be the set of all nontrivial paths.

The optimal solution is the path, $\sigma^*$, that minimizes a chosen cost function, $s : \Sigma \to \mathbb{R}_{\geq 0}$, while connecting $x_{\text{start}}$ to any $x_{\text{goal}} \in X_{\text{goal}}$ through free space,
%
\begin{align}
\sigma^* = \arg \min_{\sigma \in \Sigma} \{s(\sigma) \mid \sigma(0) &= x_{\text{start}}, \nonumber \sigma(1) \in X_{\text{goal}}, \\ & \forall t \in [0, 1], \sigma(t) \in X_{\text{free}} \}
\end{align}
where $\mathbb{R}_{\geq 0}$ is the set of non-negative real numbers. BIT* attempts to solve this problem as follows. A Random Geometric Graph (RGG) with \textit{implicit} edges is defined by uniformly sampling the free space around a start and goal position. An \textit{explicit} tree is constructed from the starting point to the goal using a heuristically guided search through the set of samples. Given a starting state $\mathrm{\mathbf{x}}_\mathrm{start}$ and goal state $\mathrm{\mathbf{x}}_\mathrm{goal}$, the function $\hat{f}(\mathrm{\mathbf{x}})$ represents an admissible estimate (i.e., a lower bound) for the cost of the path from $\mathrm{\mathbf{x}}_\mathrm{start}$ to $\mathrm{\mathbf{x}}_\mathrm{goal}$, constrained through $\mathrm{\mathbf{x}} \in \mathrm{X}$. 

Admissible estimates of the cost-to-come and cost-to-go to a state $\mathrm{\mathbf{x}} \in \mathrm{X}$ are given by $\hat{g}(\mathrm{\mathbf{x}})$ and $\hat{h}(\mathrm{\mathbf{x}})$, respectively, such that $\hat{f}(\mathrm{\mathbf{x}})$ = $\hat{g}(\mathrm{\mathbf{x}})$ + $\hat{h}(\mathrm{\mathbf{x}})$. Similarly, an admissible estimate for the cost of creating an edge between states $\mathrm{\mathbf{x}}, \mathrm{\mathbf{y}} \in \mathrm{X}$ is given by $\hat{c}(\mathrm{\mathbf{x}}, \mathrm{\mathbf{y}})$. Together, BIT* uses these heuristics to process and filter the samples in the queue based on their ability to improve the current path solution. The tree only stores collision-free edges and continues to expand until either a solution is found or the samples are depleted.

A new batch begins by adding more samples to construct a denser RGG. Had a valid solution been found in the previous batch, the samples added are limited to the subproblem that could contain a better solution.  Given an initial solution cost, $c_{\mathrm{best}}$, we can define a subset of states, $X_{\hat{f}} := \{\mathrm{\mathbf{x}} \in \mathrm{X} \Big| \hat{f}(\mathrm{\mathbf{x}}) \leq c_{\mathrm{best}} \}$ that have the possibility of improving the solution. When the metric for the edge-cost computation is Euclidean distance, the region defined by $\hat{f}(\mathrm{\mathbf{x}}) \leq c_{\mathrm{best}}$ is that of a prolate hyperspheroid with transverse diameter $c_{\mathrm{best}}$, conjugate diameter $\sqrt{c_{\mathrm{best}^2} + c_{\mathrm{min}^2}}$, and focii at $\mathrm{\mathbf{x}}_\mathrm{start}$ and $\mathrm{\mathbf{x}}_\mathrm{goal}$ \cite{Gammell2014}. 

The original implementation of BIT* is designed for shortest-distance point-to-point planning and is not customized for path following. In the remaining sections, we describe two modifications to adapt a generic sample-based  planner for the problem structure outlined in Section I.

\subsection{Curvilinear Coordinates}

Our first extension adds natural path following by using an orthogonal curvilinear planning domain \cite{Barfoot2004}. A reference path is composed of a set of discrete three Degree of Freedom (DOF) poses $P = \{\mathbf{x}_{\mathrm{start}}, \, \mathbf{x}_1,\, \mathbf{x}_2, \hdots \, ,\, \mathbf{x}_{\mathrm{goal}}\}$ with $\mathbf{x} = (x, y, \psi)$ describing the Euclidean position and yaw. We define a curvilinear coordinate, $(p,q)$, representation of the path such that the $p$-axis, $p \in [0, p_\mathrm{len}]$ describes the longitudinal distance along the path, and the $q$-axis, $q \in [q_\mathrm{min}, q_\mathrm{max}]$, is the lateral distance perpendicular to each point $p$ on the path. $q_\mathrm{min}$ and $q_\mathrm{max}$  describe the lateral place-dependant bounds of the curvilinear space at each segment of the path.

A change in distance between subsequent poses, $\Delta p$, is computed as
\begin{equation} \label{eq:one}
\Delta p = \sqrt{\Delta x^2 + \Delta y^2 + a \Delta \psi^2}.
\end{equation}

\noindent An aggregated $p$ value is stored for each discrete pose in a preprocessing step up to the total length of the path, $p_\mathrm{len}$. It is important to note that as part of \eqref{eq:one}, we incorporate a small term for changes in yaw along the repeat path tuned by a constant parameter $a$. This allows us to avoid singularities in the curvilinear coordinate space in the event of rotations on the spot by distinguishing between poses with identical positions but changing orientations. Intuitively, we set $a=1$ to balance the yaw components of the path with changes to the Euclidean axes. Increasing this value $\geq1$ expands the length of the configuration space over areas of higher curvature in the reference path, making the planner more sensitive (increasing the sampling resolution) over these portions of the path at the cost of a small amount of compute time for the increased size of configuration space. In our initial experiments, we found no appreciable benefit to changing this value from 1.

\begin{figure*}[t]
	\centering
	\includegraphics[scale=0.86]{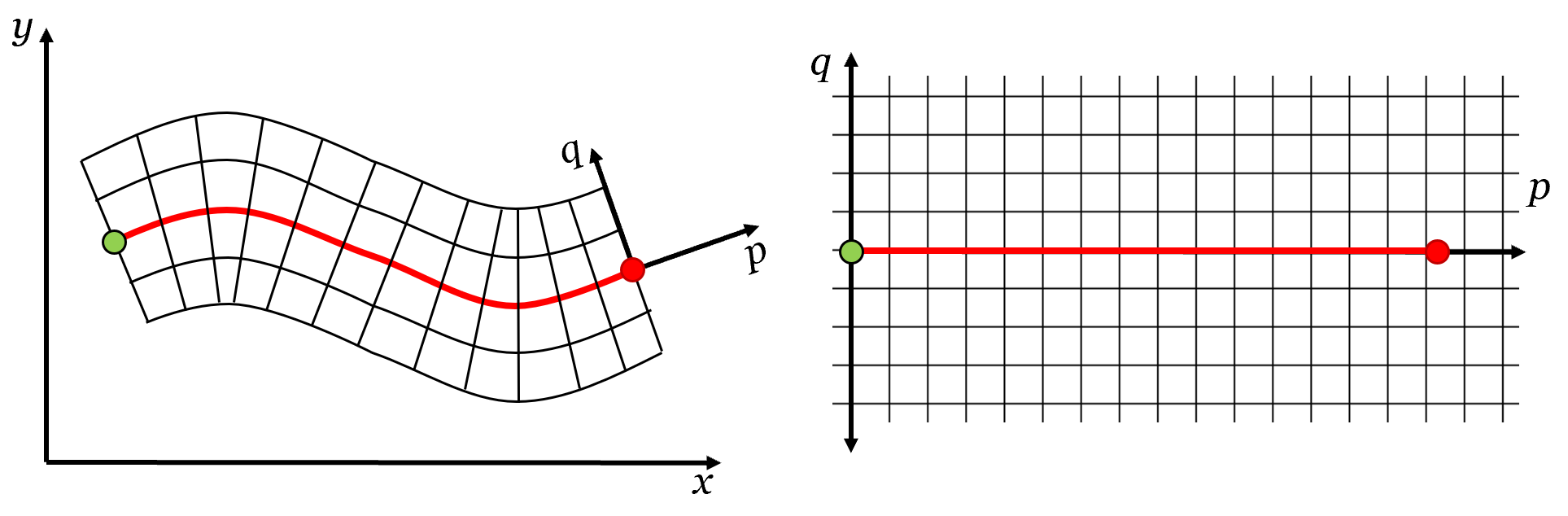}
	\caption{Left: A reference path in Euclidean coordinates shown in green, with the longitudinal and lateral components extended in a grid. Right: The corresponding representation of the path in curvilinear coordinates.}
	\label{figure3}
	\vspace{-1em}
\end{figure*}

A key observation from this definition is that all paths in Euclidean space become straight lines in $(p,q)$ space, automatically handling paths that self-intersect without requiring waypoints. By storing the $p$ values associated with each Euclidean pose from the reference path, we can uniquely map an arbitrary curvilinear point $(p_i,q_i)$ to its corresponding Euclidean point $(x_i, y_i)$ by interpolating to find a pose on the reference path closest to the target point and applying some basic trigonometry. While a small change, our novel incorporation of a heading component in the generation of the curvilinear space is critical to allow us to compute a unique map between $(p,q)$ and Euclidean space.

Generally, the inverse map cannot be guaranteed to be available due to singularities. This proves to be problematic when considering the collision checking of obstacles. Our solution is intuitive: we run BIT* in $(p,q)$ space as normal, and perform all collision checks in Euclidean space by discretizing edges and mapping the individual points back to Euclidean space for query with the obstacle costmaps. After planning a successful path in $(p,q)$ space, we use this same unique map to convert the plan back to Euclidean space for tracking with the controller.

\begin{figure*}[t]
	\centering
	\includegraphics[scale=0.8]{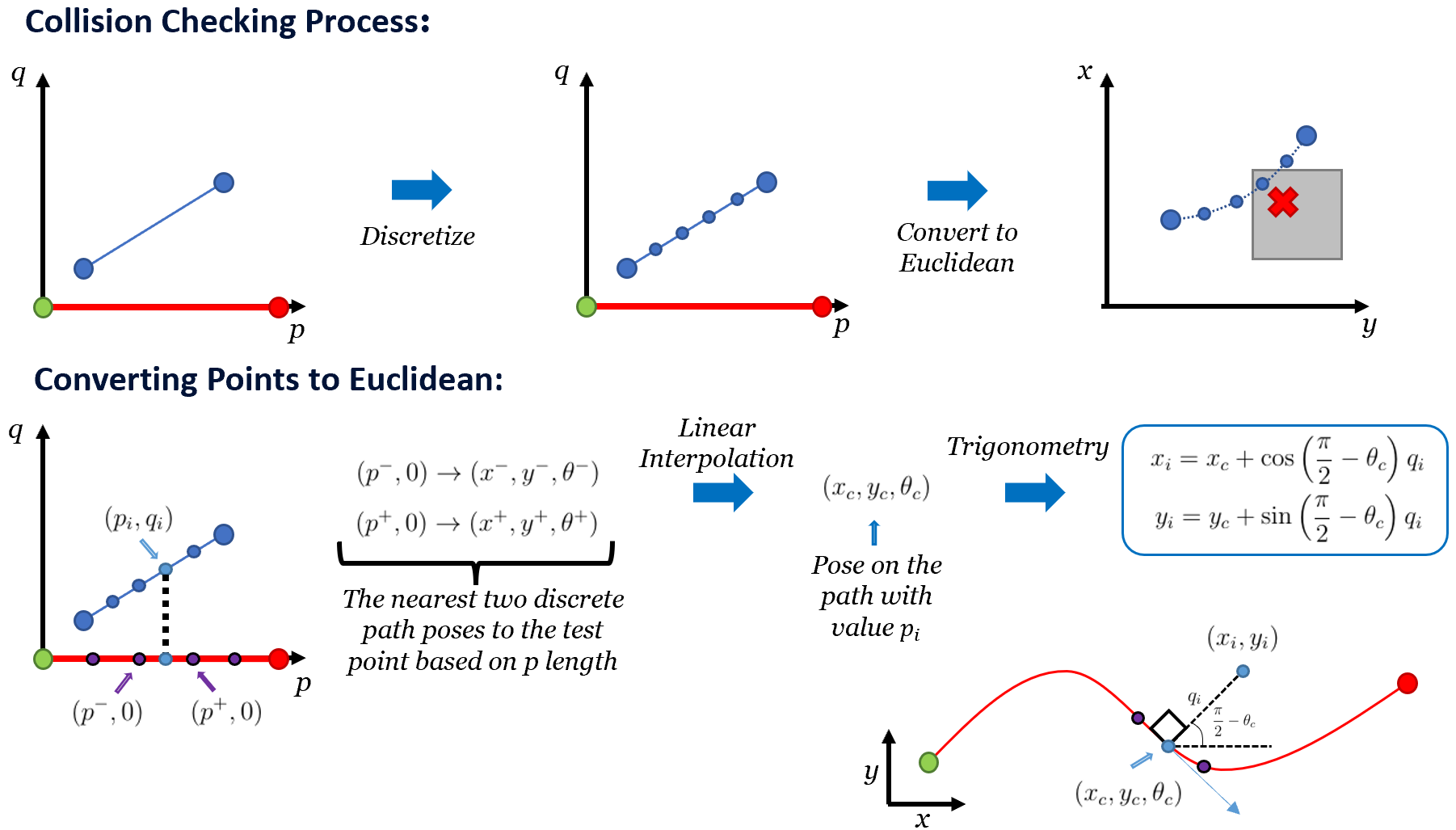}
	\caption[Collision Checking Strategy]{An illustration of the proposed collision checking scheme using curvilinear coordinates. Given an edge in curvilinear space, we can perform a collision check by finely discretizing the line into points and converting each point to Euclidean space using some basic interpolation and trigonometry. We then perform a check to see if any of the points are located inside the current obstacle map, and return a Boolean that indicates if the edge should be created in curvilinear space.}
	\label{Curve_To_Euclid}
	\vspace{-1em}
\end{figure*}

The full process for collision checking and converting points between the curvilinear planning domain and Euclidean space using the unique map is shown in Fig. \ref{Curve_To_Euclid}.

Our use of curvilinear coordinates for the planner configuration space provides several important advantages for path-following applications. It is obvious that the shortest-distance path solution in $(p,q)$ space will be the horizontal line connecting $(0,0)$ to $(p_\mathrm{goal},0)$. As we know the form of this nominal solution in advance we can generate a small subset of preseeded samples in \ac{BIT*} during each batch on the $p$-axis. This has the effect of accelerating the \ac{BIT*} convergence, as if there are no obstacles, \ac{BIT*} will immediately find the nominal solution without having to process the random samples in the space.

A final positive by-product of the curvilinear configuration space is the fact that straight-line edges in $(p,q)$ space created by a sample-based planner will be mapped back to smooth curves in Euclidean space. This is due to the fact that all plans in curvilinear space are projected relative to the underlying reference path which itself is likely to be smooth when the robot is driven with some dynamic constraints such as a minimum turning radius. This implies that after converting the \ac{BIT*} result to Euclidean space, we will have increased the \textit{smoothness} of our solutions compared to Euclidean planners.

\subsection{Weighted Euclidean Edge Metric}


Consider the 2D Euclidean planning problem where the teach path is composed of a path connecting $\mathbf{x}_\mathrm{start}$ to $\mathbf{x}_\mathrm{goal}$. The usual cost of an edge connecting two arbitrary points in space $(x_1,y_1)$ to $(x_2, y_2)$ can be expressed generally as the length $c_{21}$:

\begin{equation} \label{eq:five}
c_{21} = \int_{x_1}^{x_2} \sqrt{1 + \Big(\frac{dy}{dx}\Big)^2} dx.
\end{equation}

\noindent For our method, we incorporate an additional coefficient such that the cost of an edge increases as the lateral $y$ deviation over the length of the edge grows, scaled by a tuning parameter $\alpha$:
\begin{equation} \label{eq:seven}
c_{21} = \int_{x_1}^{x_2}(1+ \alpha y^2) \sqrt{1 + \Big(\frac{dy}{dx}\Big)^2} dx.
\end{equation}

\noindent For a straight line, this integral becomes
\begin{align} \label{eq:eight}
&c_{21} = \Bigg(1 + \frac{\alpha(y_2^3 - y_1^{3})}{3(y_2 - y_1)}\Bigg) \sqrt{(x_2 - x_1)^2 + (y_2 - y_1)^2}.
\end{align}

\noindent As $\Delta y$ approaches zero (horizontal edges) we have
\begin{align} \label{eq:nine}
\begin{aligned}
&\lim\limits_{\Delta y \rightarrow 0}
\Bigg(1 + \frac{\alpha(y_2^3 - y_1^{3})}{3(y_2 - y_1)}\Bigg) \sqrt{(x_2 - x_1)^2 + (y_2 - y_1)^2} \\
& \qquad \qquad \qquad \qquad \qquad \qquad \,= (1 + \alpha y^2) |x_2 - x_1|,
\end{aligned}
\end{align}

\noindent where $y_1 = y_2 = y$.

In \eqref{eq:eight} we obtain the Euclidean distance metric scaled by a coefficient to apply a penalty for lateral path deviation.

While this edge metric works in this simple example of a straight-line path, the result is difficult to generalize when considering arbitrarily complex reference paths in Euclidean space. In curvilinear space, however, all reference paths become horizontal lines on the $p$-axis, allowing us to directly apply this idea for the edge-cost metric in BIT* by letting $y_i \rightarrow q_i$ and $x_i \rightarrow p_i$, respectively, in \eqref{eq:eight}.

\begin{figure}[t]
	\centering
	\includegraphics[scale=0.435]{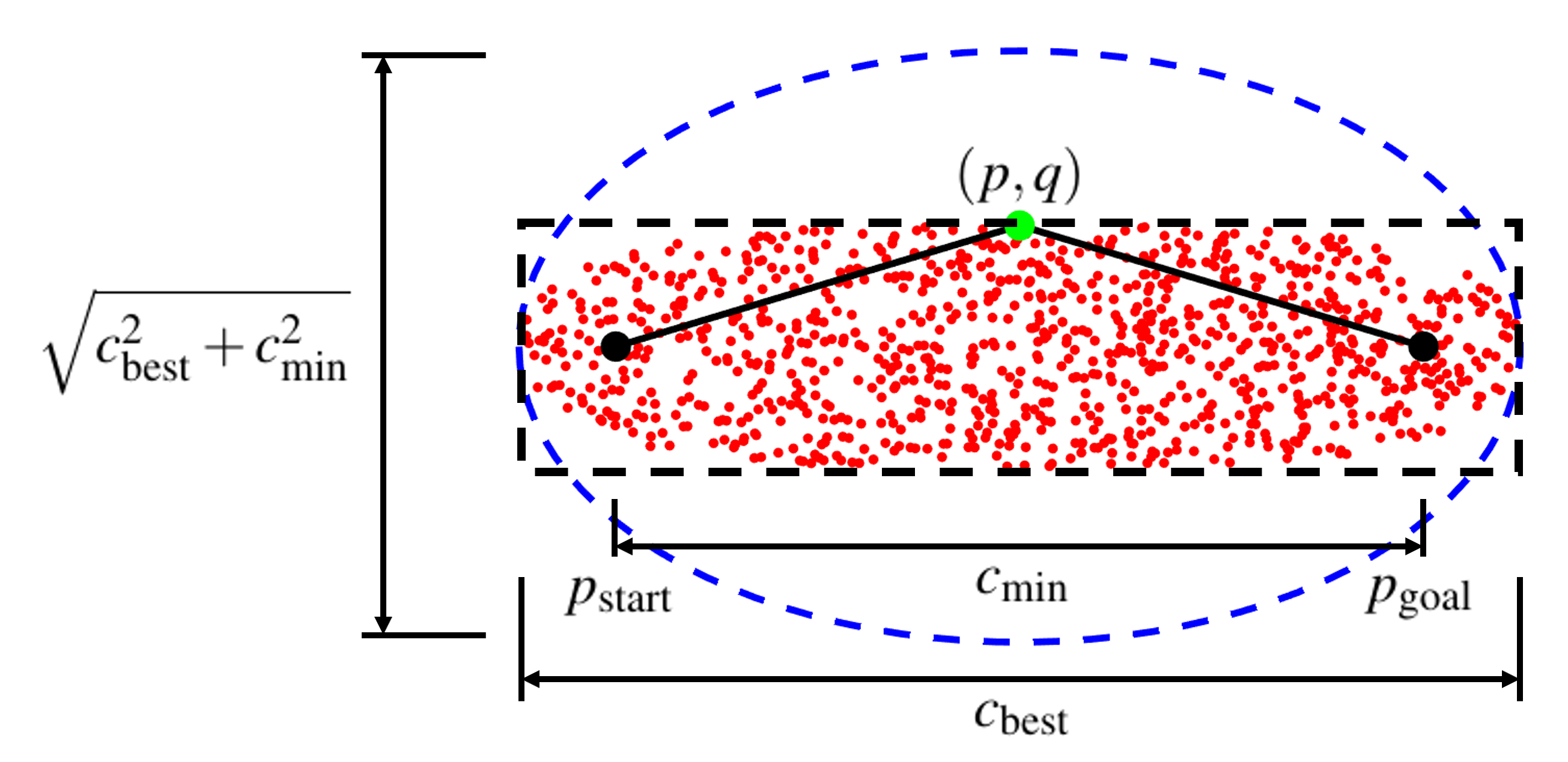}
	\caption{The informed sampling domain, $\bm{X_{\hat{f}}}$, for the Euclidean distance edge metric (blue ellipse), and the conservatively bounded laterally weighted edge metric (black box) for $\bm{\alpha = 0.5}$. Samples shown in red, were populated using rejection sampling and illustrate the true `eye-ball' distribution of the informed sampling region. As $\bm{\alpha}$ tends to zero, the domains coincide.}
	\label{figure4}
	\vspace{-1.5em}
\end{figure}

Before using this new metric in BIT*, we must first evaluate the influence of the edge-cost to the informed sampling region that constrains the RGG sub-problem following an initial solution. Consider the estimated total cost, $\hat{f}(\mathrm{\mathbf{x}}) = \hat{g}(\mathrm{\mathbf{x}}) + \hat{h}(\mathrm{\mathbf{x}})$, to incorporate an arbitrary sample, $\mathbf{x} = (p,q)$, into the path solution in curvilinear coordinates as in Fig. \ref{figure4}.

The cost is
\begin{align} \label{eq:ten}
\scalebox{0.93}{
	$\begin{aligned} [t]
	\hat{f}(\mathrm{\mathbf{x}}) = \Big(1 +& \frac{\alpha}{3} q^2\Big)\\ 
	&\Big(\sqrt{(p - p_\mathrm{start})^2 + q^2} + 	\sqrt{(p - p_\mathrm{goal})^2 + q^2} \,\Big).
	\end{aligned}$
}
\end{align}


\noindent If $\mathbf{x}$ is to improve the quality of a current solution cost, $c_\mathrm{best}$,
we require that $\hat{f}(\mathrm{\mathbf{x}}) \leq c_\mathrm{best}$. Rearranging the inequality, we have
\begin{align} \label{eq:eleven}
\begin{aligned}
&\Big(\sqrt{(p - p_\mathrm{start})^2 + q^2} + 	\sqrt{(p - p_\mathrm{goal})^2 + q^2} \,\Big) \\
& \qquad \qquad \qquad \qquad \qquad \qquad \quad \leq \frac{c_\mathrm{best}}{\Big(1 + \frac{\alpha}{3} q^2\Big)} \leq c_\mathrm{best}.
\end{aligned}
\end{align}

\noindent We note that the lateral scaling factor is always $\geq 1$ for all $q$ and that once again the left-hand term is simply the Euclidean-distance edge metric. This result implies that, conservatively, we could sample from within the informed ellipse defined by the usual Euclidean edge cost (no lateral penalty) as done by Gammell et al. \cite{Gammell2014}, and the probabilistic path convergence guarantees will remain satisfied. 

On the other hand, \eqref{eq:eleven} indicates that the lateral penalty causes our true informed sampling region to become a denser subset within the original ellipse. While difficult to describe geometrically, we can visualize this region by randomly sampling from within the outer ellipse and rejecting points with heuristic costs, $\hat{f}(\mathrm{\mathbf{x}})$, larger than the bounds. As shown in red in Fig. \ref{figure4}, we see the laterally weighted informed sampling region takes the shape of an `eye-ball' and for this problem has significantly smaller volume than the Euclidean distance ellipse. In practice, direct sampling from the eye-ball region is difficult and rejection sampling can be inefficient. However, it is possible to calculate the height of a conservative rectangle to bound the true informed sampling region and perform direct sampling from within the bounding box.



By considering a sample on the boundary at the mid-point between the start and goal, $p_\mathrm{eye} = \frac{p_\mathrm{start} + p_\mathrm{goal}}{2}$, we can compute the maximum height of the eye-ball, $q_\mathrm{eye}$, by exploiting the fact that the cost-to-come to this point, $\hat{g}(\mathrm{\mathbf{x_\mathrm{eye}}})$, is exactly equal to $\frac{c_\mathrm{best}}{2}$, where 
\begin{equation} \label{eq:3_8}
\Big(\frac{c_\mathrm{best}}{2}\Big)^2 = \Big(1 + \frac{\alpha}{3} q_{\mathrm{eye}}^2\Big)^2
\Big(\Big({\frac{p_\mathrm{start} + p_\mathrm{goal}}{2}\Big)^2 + q_{\mathrm{eye}}^2}\Big).
\end{equation}


\subsection{Other Considerations}

In an obstacle-free environment, it is true that BIT* will converge to the taught path in curvilinear space in a relatively short amount of time. The limiting factor is the generation of random continuous samples in the configuration space that fall exactly on the $q = 0$ axis. We can expedite this process significantly by initializing BIT* with a number of prepopulated samples on this axis. This allows BIT* to immediately find the optimal lowest-cost solution that follows the taught path exactly during the first batch in a matter of milliseconds, even for paths of significant length (in excess of kilometers).

This modification also drives our second integration change. In the presence of obstacles, our focus and detection capability are primarily limited to the robot's immediate surroundings. Therefore, sampling the configuration space within a sliding window, spanning from the robot's current state to the upcoming path, is sufficient for finding obstacle collisions. By exploiting this insight, we can leverage the preseeded samples to promptly generate solutions for the entire path, while accounting for local obstacles. Consequently, the number of samples required to find a solution is significantly reduced.

Our final modification takes advantage of the sliding window implementation by adopting a more efficient planning direction: from the end of the path to the current robot state, rather than the conventional approach of planning from the robot towards the goal. This decision is based on the observation that the section of the path between the end of the path and the edge of our sliding window typically represents the largest portion of the planning tree. When replanning becomes necessary due to the appearance of new obstacles, we can save valuable time by restoring only the relevant segments of the tree within the sliding window.

\section{Curvilinear Singularities and Corner Cases}

Throughout this work, we have been operating under the assumption that converting continuous paths in curvilinear coordinates to Euclidean space is a seamless process, free from singularities and discontinuities. In the vast majority of scenarios, this assumption holds true. However, there are situations, particularly during sharp inside turns, where certain regions in the curvilinear configuration space can introduce problematic singularities, challenging the validity of our assumption. In this section, we delve into a detailed exploration of how we can adapt our planning algorithm to address this issue effectively. It is worth noting that this problem is well known in curvilinear coordinate representations such as Fren\'et Frames, but is often ignored in the literature. Our method of addressing this issue is another key contribution of this work.

\subsection{Singularity Regions}

\begin{figure}[t]
	\centering
	\captionsetup[subfloat]{labelfont=scriptsize,textfont=scriptsize}
	\subfloat[Curvilinear plan with constant offset.]{\includegraphics[scale=0.47]{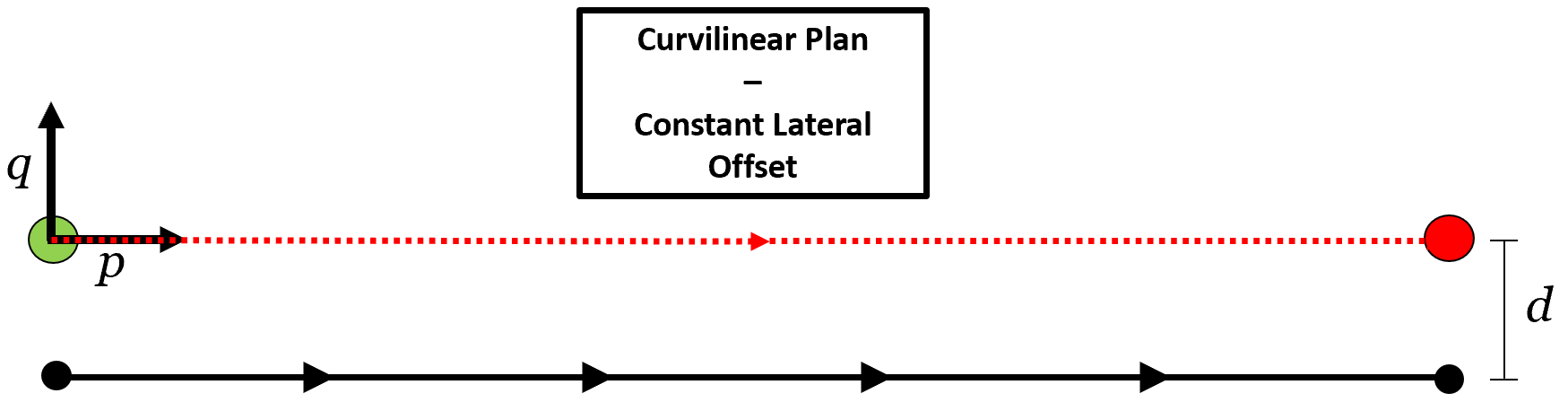}\label{fig:Singularity_Toy_Problem1}}
	
	\subfloat[Corresponding Euclidean reference path and planner result after conversion to Euclidean space.]{\includegraphics[scale=0.53]{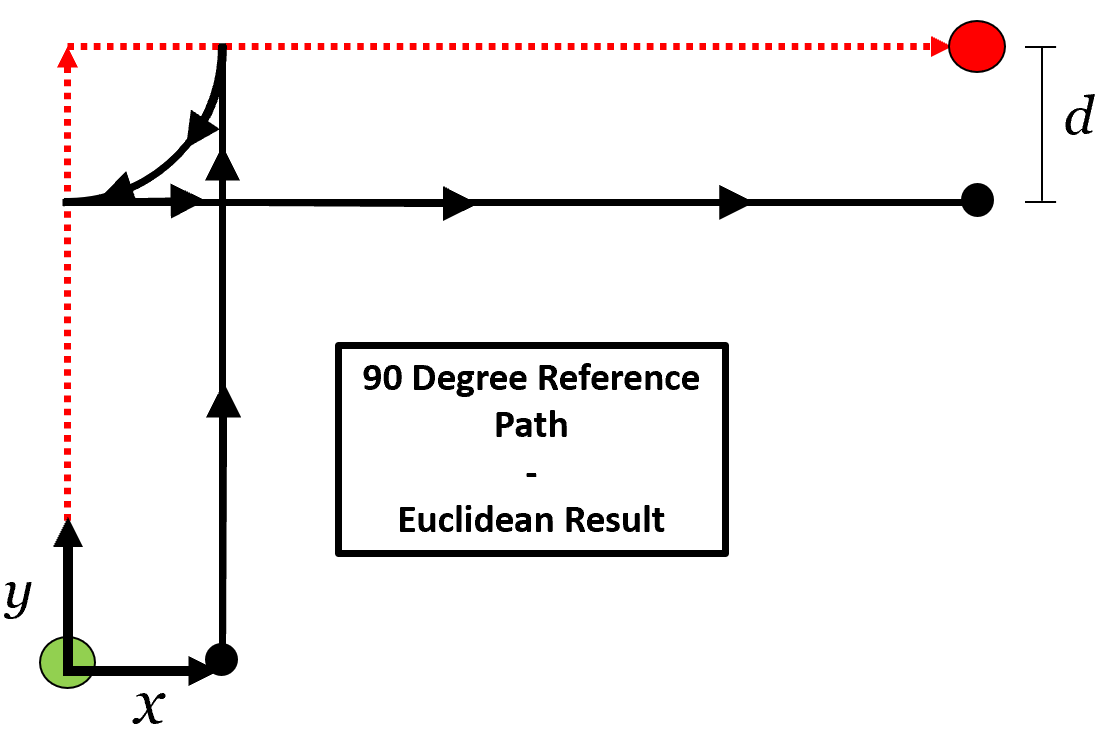}\label{fig:Singularity_Toy_Problem2}}
	
	\caption[Corner Case Toy Problem]{In this figure we illustrate a toy problem to help conceptualize the curvilinear space singularity issue. Consider a red Euclidean reference path with a sharp 90 degree turn and its curvilinear straight line representation. In curvilinear space, if we wished to follow the reference path with a constant lateral offset of $\bm{d}$ [m] we simply generate the black horizontal path solution in (a). After being converted to Euclidean space in (b), we find that taking this path results in an undesirable three-point turning behaviour.}
	\label{Singularity_Toy_Problem}
	\vspace{-2.5em}
\end{figure}

To better understand the singularity problem, let us consider the toy problem depicted in Fig. \ref{Singularity_Toy_Problem}. Imagine a reference path in Euclidean space that consists of a straight-line path that takes a sharp 90-degree turn to the right halfway through. In curvilinear representation, this path becomes a straight horizontal line. Now, if we attempt to follow a line with a fixed lateral offset from this initial path in curvilinear space and convert it back to Euclidean space, we will end up with the irregular path shown in Fig. \ref{fig:Singularity_Toy_Problem2}.

\begin{figure}[t]
	\centering
	\includegraphics[scale=1.55]{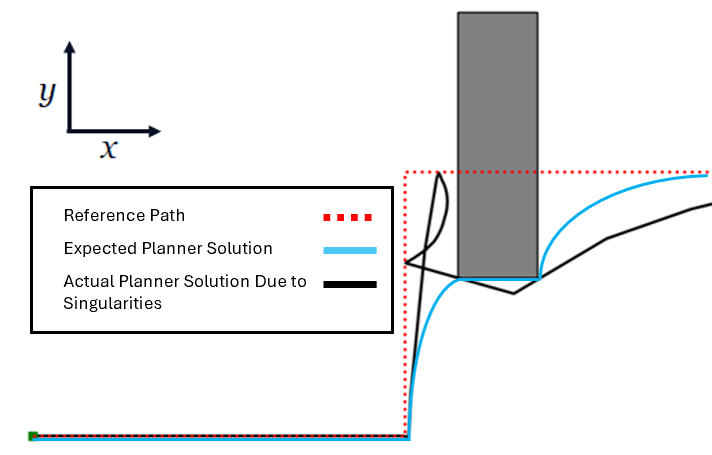}
	\caption[The Effect of Curvilinear Singularity Regions]{In this scenario, we have another sharply turning reference path. The presence of the obstacle near to the corner necessitates the planner to cross a singularity region resulting in the odd Euclidean path result shown in black. We would prefer to take a path shown in blue.}
	\label{Inside_Singularity_Example}
	\vspace{-1.5em}
\end{figure}

While this path is technically valid, a maneuver such as this is generally not desirable. This is an extreme example. However, generally these singularities will always be present the further a robot deviates laterally from the reference path. When the reference path is smooth with low curvature, singularities appear far enough away that they can be ignored, as is often done in self-driving applications using road centre lines. The occurrence of this `three-point-turn' behaviour arises when the planned path in BIT* traverses the curvilinear space with a non-zero lateral $q$ component over regions where the reference Euclidean path exhibits excessive inside curvature. A common scenario where this can occur is when an obstacle is placed near the inside corner of a sharp turn, as illustrated in Fig. \ref{Inside_Singularity_Example}.

In this instance, a sharp turn in the reference path combined with an obstacle located very close to this curve results in the irregular `three-point-turn' when the curvilinear path solution is converted back to Euclidean space. Although this path is technically valid, such behaviour is generally not desirable. The occurrence of this behaviour arises when the planned path in BIT* traverses the curvilinear space with a nonzero lateral $q$ component over regions where the reference Euclidean path exhibits excessive inside curvature.

Although excessive curvature is typically manually avoided by the operator during the teach phase of \ac{VTR3} \cite{Wu2022} due to the wear imposed on a large differential-drive robot and limitations with localization accuracy, it is still important to address this issue comprehensively. Our proposed solution involves identifying these problematic regions during an offline precomputation step. Then, during runtime, we modify our planning algorithm to naturally avoid these regions.

To define the singularity regions, we consider an arbitrary point in curvilinear space, denoted as $(p,q)$, with the corresponding projected Euclidean space point as $(x_p, y_p)$. The singularity regions can be described completely by the inequality

\begin{equation}  \label{eq:3_9}
q_{\mathrm{max}}(p) \ge q > q_{\mathrm{min}}, 
\end{equation}

\noindent where $q_{\mathrm{max}}(p)$ represents the maximum lateral bounds of the corridor defined by the reference path at each point $p$, and $q_{\mathrm{min}}$ is defined as the distance to the nearest point on the reference path from the Euclidean point $(x_p, y_q)$. In other words, a curvilinear point falls within a singularity region if, when converted to Euclidean space, the $q$ value does not match the distance to the closest Euclidean point on the reference path.

While defining the singularity regions is a straightforward task, discretizing the configuration space to compute the interior boundary (where $q = q_{\mathrm{min}}$) is neither practical nor elegant. Instead, we can rely on another observation: these regions are closely linked to the curvature of the reference path, specifically the instantaneous \ac{ROC}. Our key insight is to notice that all points where $q > \mathrm{ROC}$ also satisfy the equality in \eqref{eq:3_9}. Although the \ac{ROC} does not fully describe the singularity regions, it provides a promising starting point for our search. By precomputing the ROC at each point $p$ along the path using the reference path, we can expand along lines of constant lateral distance and test for condition $q = q_{\mathrm{min}}$ to be satisfied. This condition is satisfied for all points outside the singularity regions and will first occur at the boundary.

\begin{figure}[t]
	\centering
	\captionsetup[subfloat]{labelfont=scriptsize,textfont=scriptsize}
	\subfloat[Euclidean reference path.]{\includegraphics[height=5.8cm]{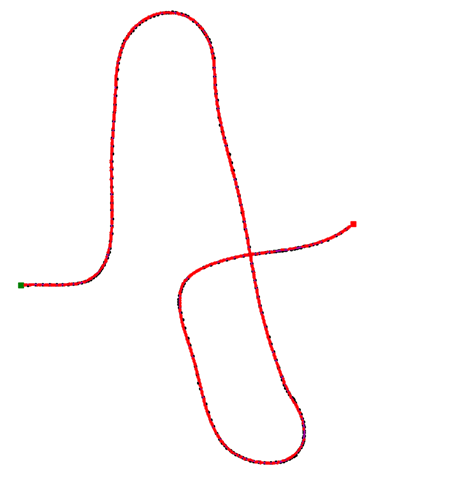}\label{fig:ROC_Example1}}
	\hfill
	\subfloat[Curvilinear path representation with instantaneous radius of curvature overlay.]{\includegraphics[height=5.5cm]{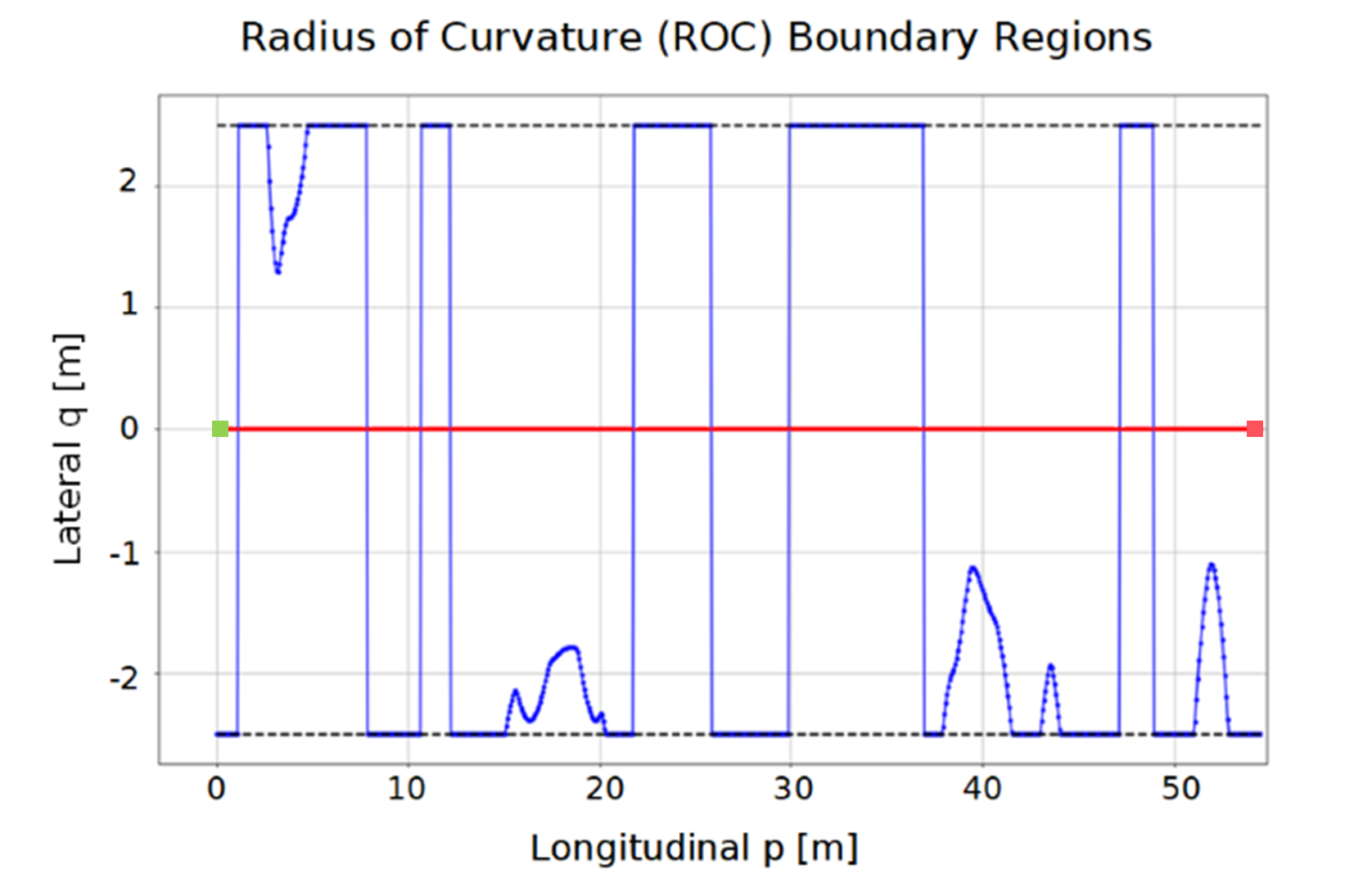}\label{fig:ROC_Example2}}
	\caption[Reference Path Instantaneous Radius of Curvature Plotting]{In this figure we show an intricate reference path in (a) and its corresponding curvilinear representation in (b). At each point along the path, we compute an instantaneous \ac{ROC}, plotting the result in blue. Note that we clamp the \ac{ROC} magnitudes at the maximum corridor bounds, $\bm{\pm q_{\mathrm{max}}}$ for readability.}
	\label{ROC_Example}
	\vspace{-1em}
\end{figure}

\begin{figure}[t]
	\centering
	\captionsetup[subfloat]{labelfont=scriptsize,textfont=scriptsize}
	\subfloat[Instantaneous \ac{ROC} in $(p,q)$ space.]{\includegraphics[height=6.7cm]{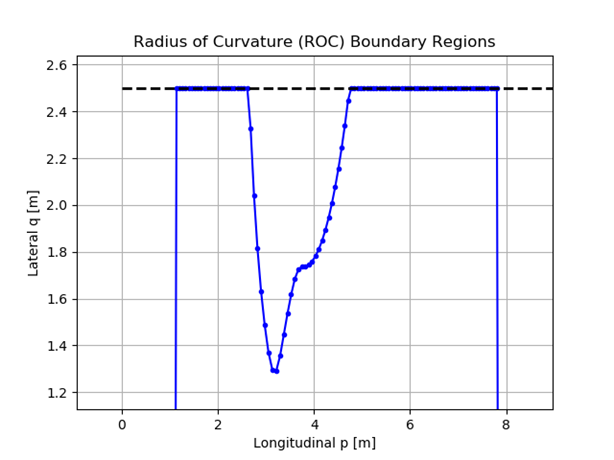}\label{fig:Singularity_From_ROC1}}
	\hfill
	\subfloat[Singularity boundary region with wormhole edges.]{\includegraphics[height=6.5cm]{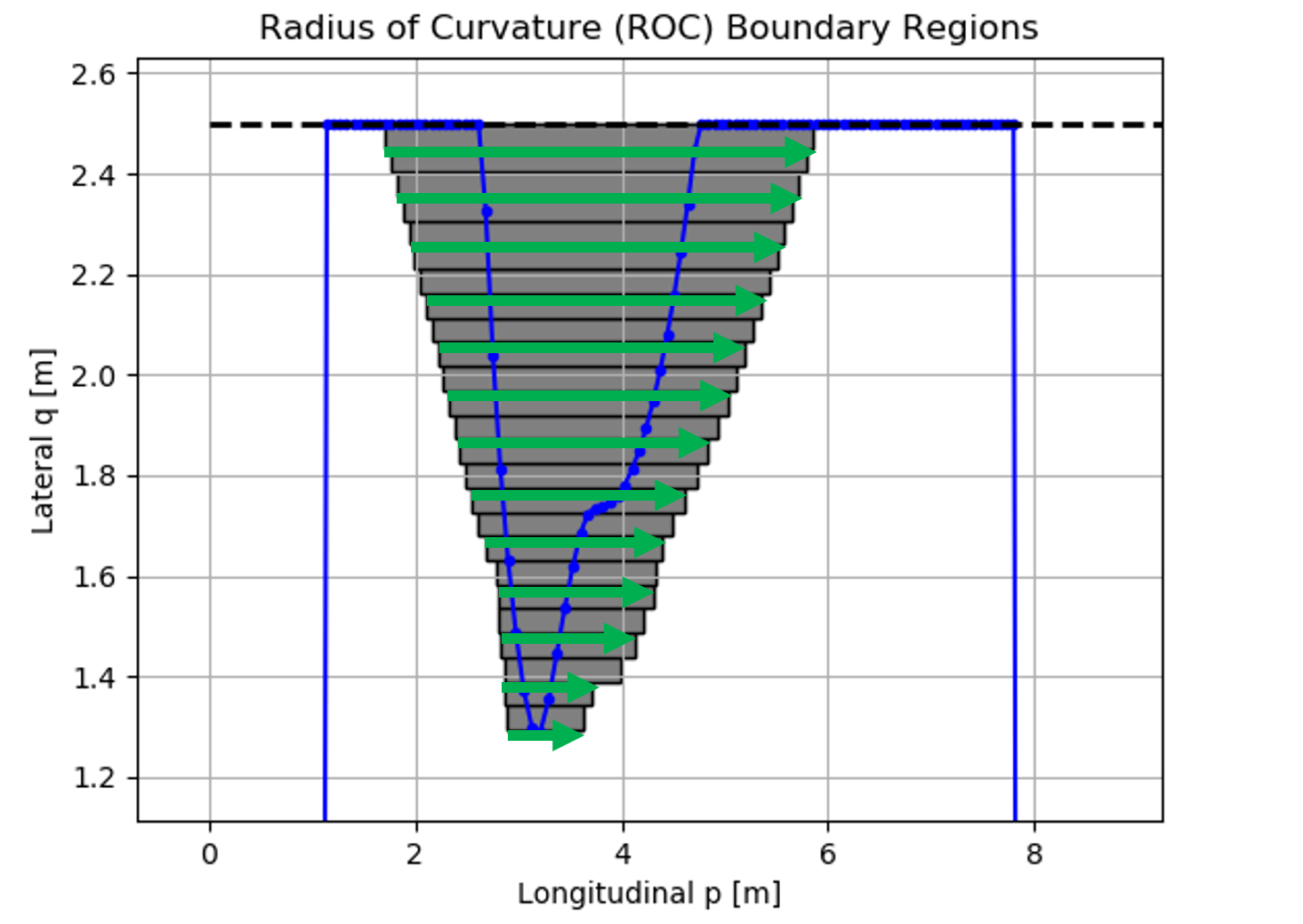}\label{fig:Singularity_From_ROC2}}
	
	\caption[Identifying Singularity Regions]{In this figure we illustrate the generation of the curvilinear singularity regions. Using the \ac{ROC} plot as a starting pot, we discretize the space laterally and expand from the \ac{ROC} plot to the left and right along the $p$-axis, testing until condition \eqref{eq:3_9} is invalidated. Eventually, we are left with the grey singularity region shown in (b). We then preseed a number of wormhole edges shown in green that BIT* can use to find valid path solutions that cross the singularities.}
	\label{Singularity_From_ROC_Example}
	\vspace{-1em}
\end{figure}

A typical plot depicting the \ac{ROC} of a reference path is illustrated in Fig. \ref{ROC_Example}, and Fig. \ref{Singularity_From_ROC_Example} demonstrates how to generate the corresponding singularity region. To ensure tractability, we discretize the lateral and longitudinal directions with a resolution of 10 cm and conservatively approximate the singularity regions using a series of adjoining rectangles. We do not bother calculating the singularity regions outside of the maximum lateral boundary for our reference path as we never intend to plan in these regions.

\subsection{Wormhole Generation}
Now that we have a better understanding of the singularity regions and how they appear in the planner, we discuss how to avoid them. One possible approach would be to treat these regions as obstacles and prevent the planner from generating paths that pass through them. While this technique does work, it would eliminate the option of taking the inside corner to circumvent an obstacle on a sharp turn, as shown in Fig. \ref{Inside_Singularity_Example}. However, it is clear in this example that taking the inside corner could still be the most efficient choice. Therefore, solely relying on this approach should be avoided if possible. 

An important insight we can exploit is that the boundary points of the singularity regions, denoted as $(p_l, q)$ and $(p_r, q)$ for the left and right boundary points of $q$, respectively, correspond to the same Euclidean position following conversion. The distinction is in their headings, which can vary significantly. This concept can be seen most clearly in Fig. \ref{fig:Singularity_Toy_Problem2}, where the robot path crosses over itself at the intersection diagonally from the reference path corner. Instead of following the three-point-turn path, it may be better to first reach this intersection and then turn on the spot 90 degrees clockwise and continue following the reference path, \textit{jumping over} the three-point-turn segment of the path.

\begin{figure*}[t]
	\centering
	\captionsetup[subfloat]{labelfont=scriptsize,textfont=scriptsize}
	\subfloat[Lateral BIT* with wormholes.]{\includegraphics[scale=0.675]{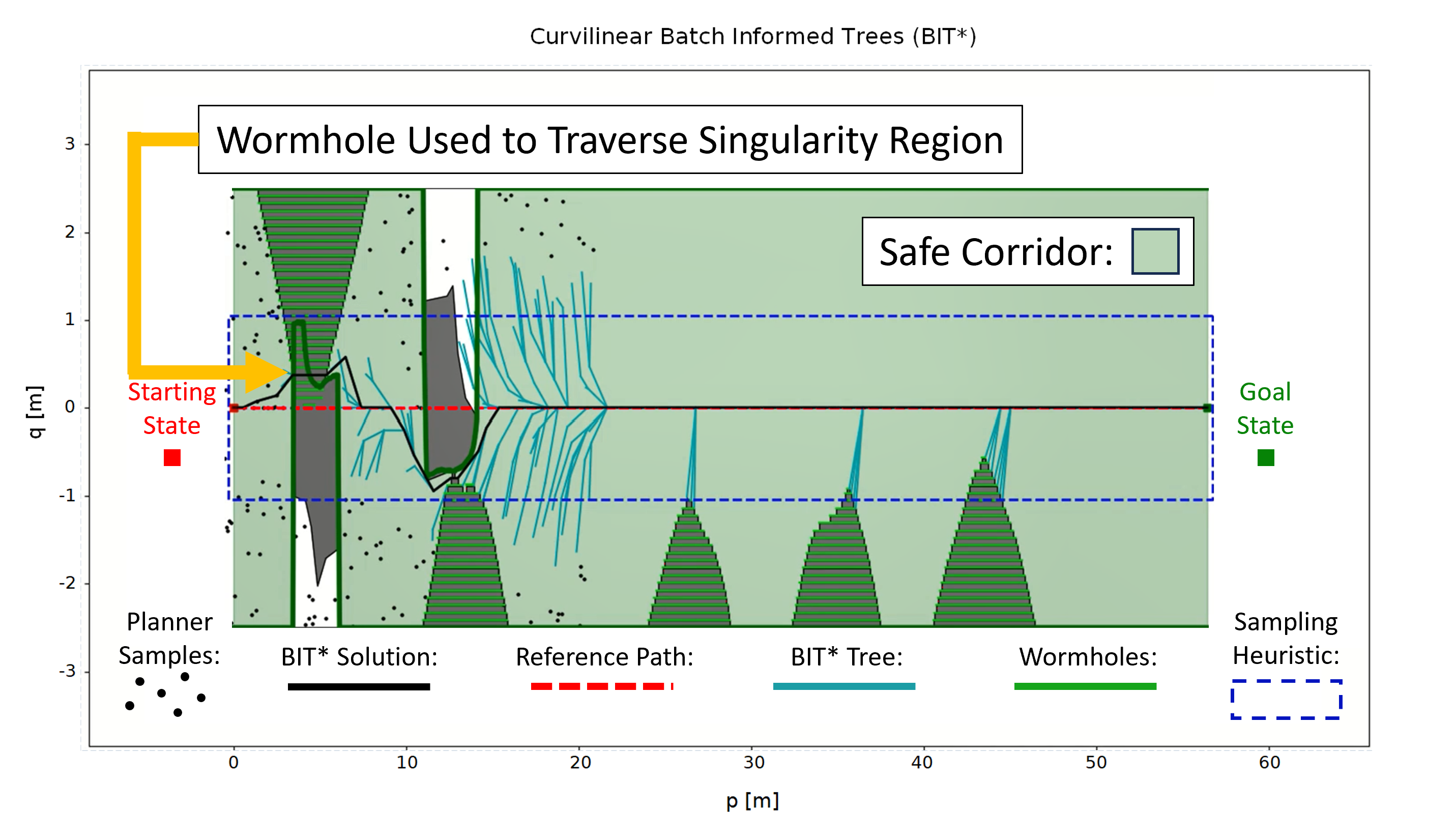}\label{fig:Wormhole_Demo1}}
	
	\subfloat[Corresponding Euclidean Plan.]{\includegraphics[scale=1.21]{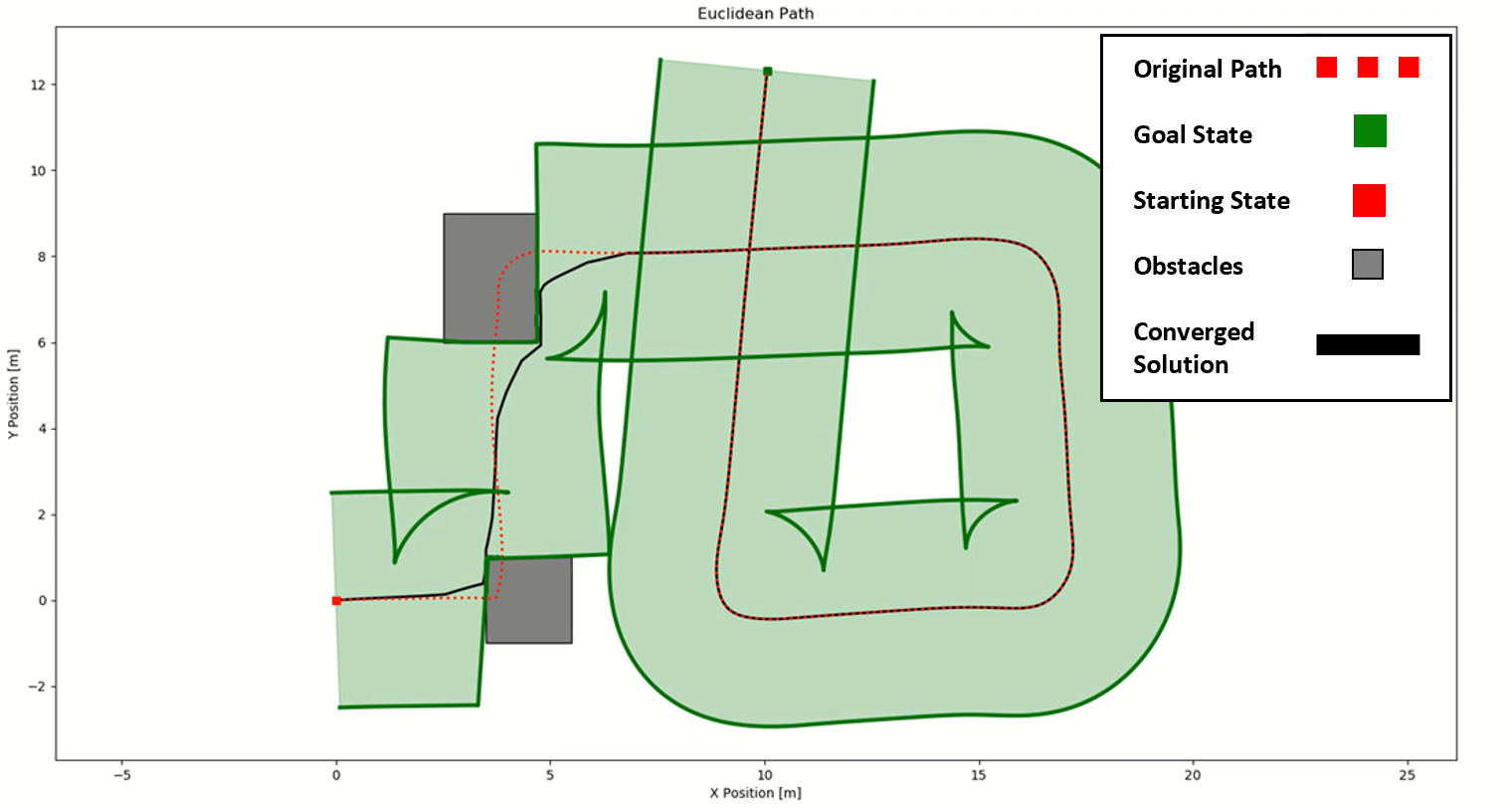}\label{fig:Wormhole_Demo2}}
	
	\caption[Constructing Singularity Avoiding Wormholes Example 2]{In this figure, we show a representative planning scenario using wormholes to find better path solutions. We find that the use of wormholes allows us to generate paths that tightly hug obstacles (grey), opting for turns on the spot to further reduce lateral path deviation. These turns on the spot allow us to generate safe-corridor constraints, (green), that the controller can use to smooth out the trajectory while maintaining collision-avoidance properties.}
	\label{Wormhole_Demo}
	\vspace{-1em}
\end{figure*}

In other words, by traversing from the leftmost boundary point to the rightmost boundary point of a singularity region, we effectively have an edge that, when followed, executes a turn on the spot in Euclidean space. This realization provided the intuition that instead of planning through the singularity regions we can try to leap past them. To achieve this, we introduce the concept of \textit{wormholes}. In this context, a wormhole refers to pairs of vertices in the curvilinear planning domain that, when connected by the planning tree, enable \textit{teleportation} across the singularity regions. We can prepopulate several wormhole edges in the planning tree that serve as passageways, opening up the possibility of traversing inside corners during planning. An illustration of the preseeded wormholes for the reference path problem in Fig. \ref{Inside_Singularity_Example} is depicted in Fig. \ref{Singularity_From_ROC_Example} and we demonstrate a planning example using wormholes in Fig. \ref{Wormhole_Demo}. In this example, our planner is able to find elegant inside corner solutions in Euclidean space avoiding the two obstacles shown in grey by using the pre-populated wormhole edges to bypass the singularities.

We do not wish to allow the planner to use these means of transporting the configuration space for free. However, as we recall taking a path through a wormhole results in a plan that turns on the spot, a generally undesirable feature. As such we can impose a cost for these edges that is proportional to the amount of rotation that would be incurred and a tunable parameter that allows the user to adjust how willing they are for the planner to take these passageways. If a small turn on the spot behaviour yields a significant path cost reduction compared to a long outside corner path, then in some cases it may be acceptable to allow this behaviour.


\section{Model Predictive Control}

\subsection{Tracking Model Predictive Control}
In this section, we will delve into the mathematical formulation of our Model Predictive Control (MPC) schemes. To ensure generality and versatility across various robotic platforms, we adopt a representation that leverages matrix Lie group theory, drawing upon notation from Barfoot \cite{Barfoot2017}. Similar to the implementation by Teng et al. \cite{Teng2023} and Chang et al. \cite{Chang2020}, we represent the robot's state using the Special Euclidean Group $\LieGroupSE{3}$\footnote{We formulate our problem generally in $\LieGroupSE{3}$. However, for planar robots, $\LieGroupSE{2}$ can be used to save a marginal amount of computing effort.}, which allows us to encapsulate both the position and orientation of the robot in a single mathematical object. This choice allows us to create an MPC framework that can seamlessly accommodate a wide range of vehicles, from 6 \ac{DOF} drones to differential-drive unmanned ground vehicles (UGVs) and beyond.


Given a robot in a moving coordinate frame $\CoordinateFrame{v}$ with respect to a global frame $\CoordinateFrame{i}$, we define the generalized 6 \ac{DOF} velocity vector $\Matrix{\varpi}$ in the moving frame as
\begin{align}
&\Matrix{\varpi} = \bbm \Matrix{v}_v^{vi} \\ \Matrix{\omega}_v^{vi} \ebm, \label{eq:4_1}
\end{align}
composed of the linear and angular velocity vectors, $\Matrix{v}$ and $\Matrix{\omega}$, respectively. The pose of the vehicle in the global frame is subsequently defined by the transformation matrix 
\begin{align}
&\T \triangleq \T_{vi} = \bbm \Matrix{C}_{vi} & \Matrix{r}_{v}^{iv} \\ \Matrix{0}^T & 1 \ebm. \label{eq:4_2}
\end{align}
By constraining the generalized velocity in \eqref{eq:4_1}, we can enforce different robot kinematic models using a selection matrix, $\Matrix{P}$, to isolate the nonzero components of $\Matrix{\varpi}$. In our specific application, we will be working with a unicycle model with only forward linear and angular yaw velocities, $v$ and $\omega$, respectively. However, it is clear to see that by specifying a different projection matrix, we can enforce alternative kinematic constraints.  We make the following substitution in terms of a lower-dimensional velocity vector $\Matrix{u}$, such that
\begin{equation}\label{eq:4_3}
\Matrix{\varpi} = \bbm v & 0 & 0 & 0 & 0 & \omega \ebm^T =
\Matrix{P}^{T} \ub,
\end{equation}
where we have:
\begin{align}
&\Matrix{P} = \bbm 1 & 0 & 0 & 0 & 0 & 0 \\
0 & 0 & 0 & 0 & 0 & 1
\ebm, \quad \Matrix{u} = \bbm v \\ \omega \ebm. \label{eq:4_4}
\end{align}

\begin{figure}[t]
	\centering
	\includegraphics[scale=0.6]{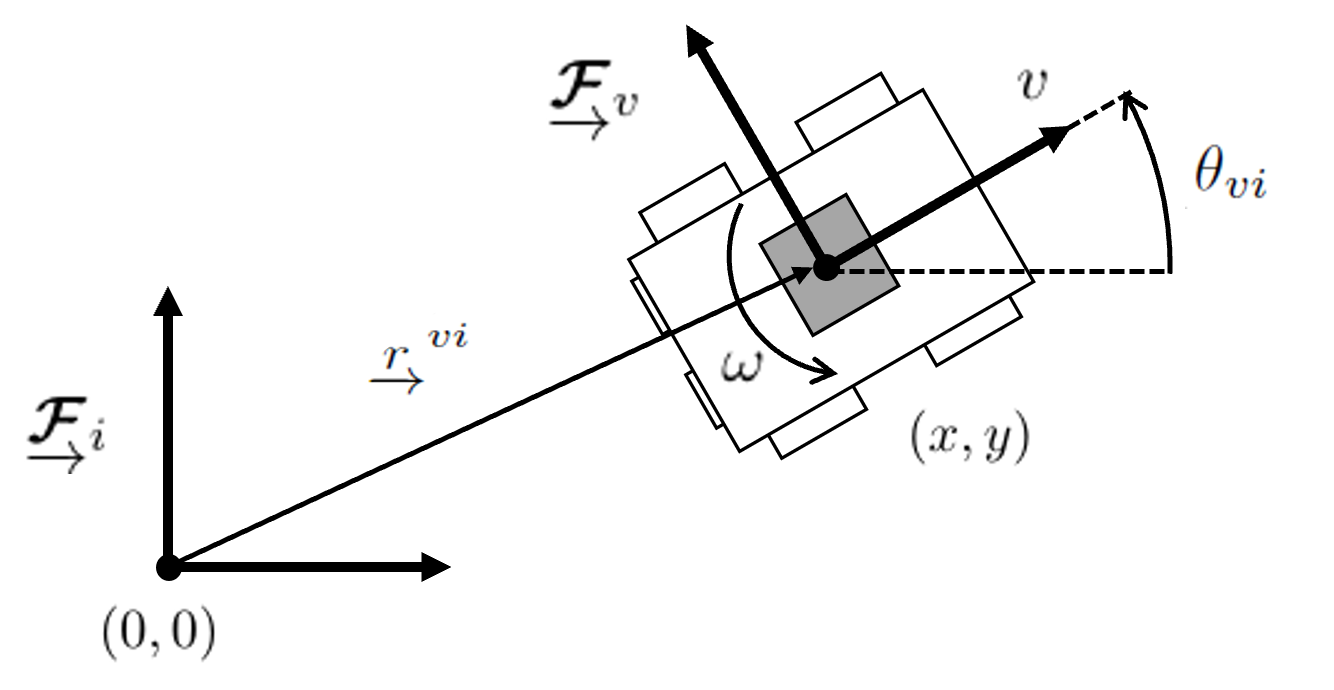}
	\caption[Robot Motion Model]{The definition of our robot state variables for a differential-drive robot.}
	\label{Unicycle_Kinematics}
	\vspace{-1em}
\end{figure}

We wish to derive a model-predictive controller to compute the optimal sequence of $k$ control inputs, out to a horizon of $K$ time steps with a period of $h$ seconds, to minimize the error between some reference path, $\Trefk$, and the predicted trajectory of the robot. This path-tracking task can be realized by solving the least-squares optimization problem:
\begin{subequations} \label{test}
	\begin{align} \label{eq:4_6a}
	&\begin{aligned} [t]
	&\mathrm{argmin} \, J(\T ,\Matrix{u}) =
	\sum_{k=1}^K 
	\ln (\Trefk \Tk^{-1})^{\vee^{T}}
	\Matrix{Q}_k
	\ln (\Trefk \Tk^{-1})^{\vee}\\
	&\qquad \qquad \qquad \qquad \qquad \qquad \qquad \qquad \qquad \quad \, \, \,+
	\uk^T \Matrix{R}_k \uk \\
	&\mathrm{s.t.} 
	\end{aligned} \\
	&\Tkone =
	\exp \Big( \Puk^\wedge h \Big) \Tk, \quad k = 1, 2, \hdots , K \label{eq:4_6b} \\
	&\Matrix{u}_{\mathrm{min},k} \leq \uk \leq \Matrix{u}_{\mathrm{max},k}, \quad k = 1, 2, \hdots , K. \label{eq:4_6c}
	\end{align}
\end{subequations}
%
%

Using Lie groups and the skew-symmetric and inverse skew-symmetric operators $\wedge$ and $\vee$ respectively, our objective function \eqref{eq:4_6a} aims to minimize the pose error between the trajectory we intend to generate and a reference trajectory, while simultaneously trying to minimize control effort. \eqref{eq:4_6b} and \eqref{eq:4_6c} are our generalized kinematic constraint and actuation limit constants, respectively.

For the direct-tracking control implementation, we obtain the reference poses at each time step $\Trefk$ directly from the current densely discretized Euclidean \ac{BIT*} solution that guides the robot from its current state to the path's endpoint. To determine the reference poses along the trajectory for each time step, we interpolate poses from the current planner solution with a separation distance denoted as $p_{\mathrm{ref},k}$. This allows us to try to maintain a nominal desired path-tracking velocity $v_{\mathrm{ref}}$ such that
\begin{align}
&p_{\mathrm{ref},k} = v_{\mathrm{ref}} h k \label{ref_pose_gen}.
\end{align}

\noindent The weights $\Matrix{Q}_k$ and $\Matrix{R}_k$ can be tailored to the specific application, allowing adjustment of the relative importance of different tracking degrees of freedom. Likewise, the actuation limits $\Matrix{u}_{\mathrm{min},k}$, $\Matrix{u}_{\mathrm{max},k}$, and $v_{\mathrm{ref}}$ are set based on the robot parameters. 
To generate the trajectory with the lowest cost, we solve this \ac{MPC} problem over a horizon of $K$ steps. Following the standard \ac{MPC} approach, we apply the first velocity command to the robot and subsequently reinitialize the problem at each control loop request based on the latest available information.


\subsection{Homotopy-Guided Corridor Model-Predictive Control}
To address the limitations of the direct-tracking \ac{MPC} approach, we propose an alternative architecture that provides a concrete guarantee on collision avoidance. Instead of blindly following the output of the path planner, which may not always be kinematically feasible, our approach involves directly tracking the driven reference path that is guaranteed to be kinematically feasible  in the VT\&R paradigm. We then incorporate the planner solution into a hard spatial constraint within the \ac{MPC} problem.

To understand our approach, it is crucial to discuss the concept of path homotopies. In an environment with obstacles between a robot and its goal, the presence of obstacles introduces different potential path-homotopy classes. The emergence of multiple homotopy classes creates local minima in the optimization problem, making it challenging to find the globally optimal solution. To tackle this challenge, we draw inspiration from the work of Linigar et al. \cite{Liniger2015}. We utilize the planner's solution to identify the most promising homotopy class among the set of possible paths. By considering the characteristics of the selected homotopy class, we define a series of lateral corridor path constraints that shape the \ac{MPC} trajectory planning within a convex planning space. These lateral corridors serve as spatial constraints that guide the \ac{MPC} toward a globally optimal solution for the trajectory-planning problem.

The new \ac{MPC} problem is initialized using the current path solution and optimized within the constrained convex space to gain three key advantages.
Firstly, utilizing the collision-free BIT* solution for corridor constraints decouples the \ac{MPC} solution from the planner, providing a guarantee on collision avoidance. Secondly, this lets us work with approximate planner solutions to define corridors, enabling safe operation at higher velocities compared to direct-tracking. Lastly, reference poses are consistently chosen from the original taught path instead of the planner's solution, potentially leading to slight improvements in path-following error.

\begin{figure}[t]
	\centering
	\includegraphics[scale=0.6]{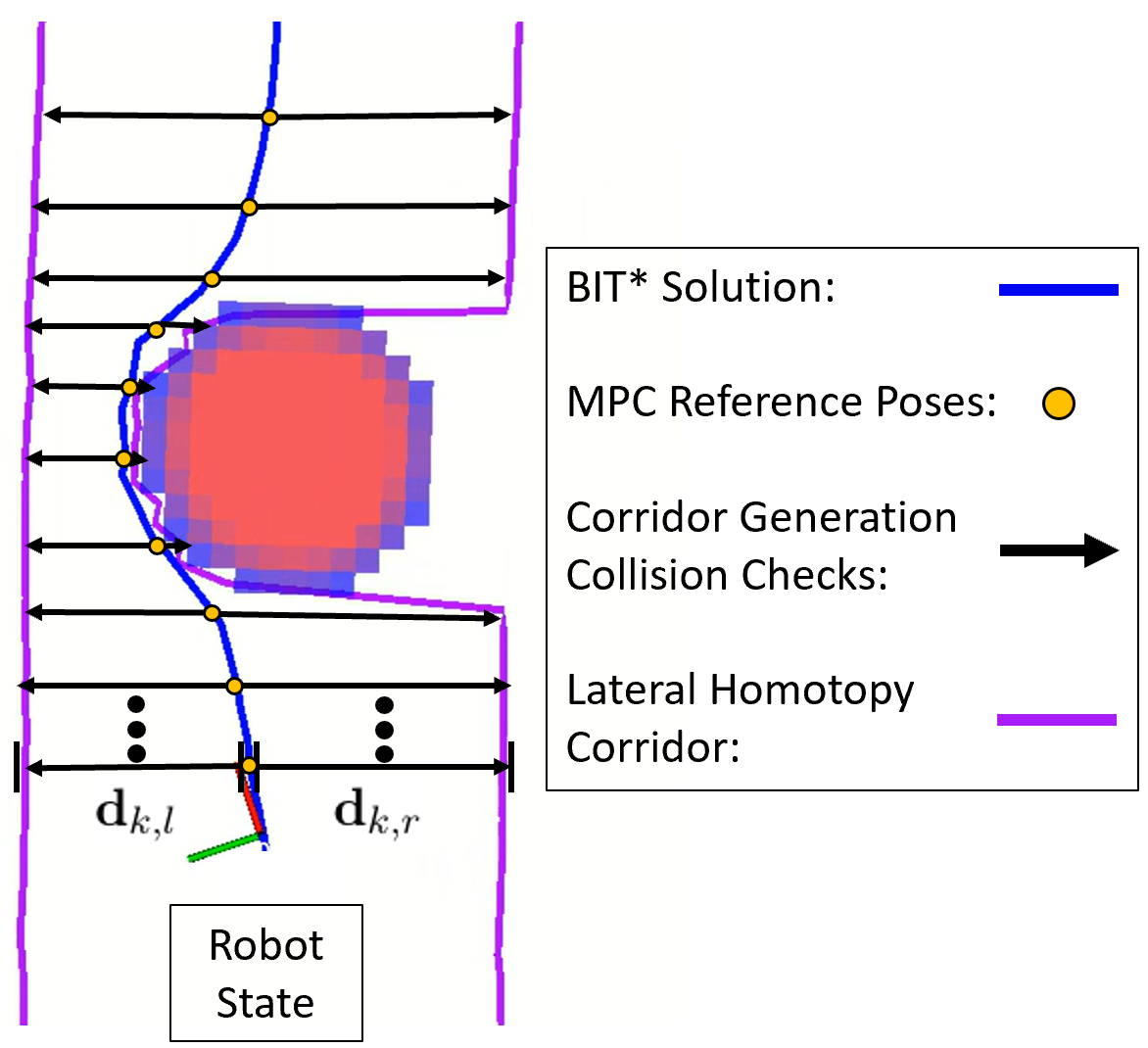}
	\caption[Lateral Corridor Generation Process]{We generate a dynamic safe lateral corridor constraint (purple) using the current planner solution (blue). Given reference poses along the path, we collision check laterally in both directions, starting from the planner solution to the maximum place-dependant corridor bounds. If there are no collisions, the corridor is set to the maximum bounds. If there is a collision, the corridor is set to the vertex immediately preceeding the collision point and used as state constraints in the homotopy-guided MPC.}
	\label{Corridor_Generation_Example}
	\vspace{-2em}
\end{figure}

The revised \ac{MPC} optimization problem is similar to \eqref{test}, but now includes an additional lateral corridor constraint:
\begin{equation}
-\Matrix{d}_{k,l} \leq \Matrix{1}_{2}^{T} \Trefk \Tk^{-1} \Matrix{1}_{4} \leq \Matrix{d}_{k,r}, \quad k = 1, 2, \hdots , K. \label{eq:4_9d}	
\end{equation}
%
Here, $\Matrix{1}_{i}$ represents the \textit{i}-th column of the identity matrix, with $\Matrix{d}_{k,l}$ and $\Matrix{d}_{k,r}$  denoting the maximum allowable lateral deviation at each point on the left and right sides of the path that define the path-homotopy corridor. The values of $\Matrix{d}_{k,l}$ and $\Matrix{d}_{k,r}$ are computed on-demand by the controller and the process is outlined as follows:

Upon obtaining the curvilinear space representation of the desired reference pose from the latest \ac{BIT*} solution, $(p_{\mathrm{ref},k}, q_{\mathrm{ref},k})$, in curvilinear coordinates, we construct two test edges extending from $(p_{\mathrm{ref},k}, q_{\mathrm{ref},k})$ to the maximum safe-corridor boundaries $(p_{\mathrm{ref},k}, q_{\max})$ and $(p_{\mathrm{ref},k}, -q_{\max})$, where $q_{\max}$ is the place-dependant maximum lateral boundary set by the user at each point along the reference path. We then perform collision checks on these edges using our standard procedure and set $\Matrix{d}_{k,l}$ and $\Matrix{d}_{k,r}$ equal to the lateral component of the last collision-free vertex along the discretized test edges. This process is illustrated in Fig. \ref{Corridor_Generation_Example}.

Just as in the direct-tracking implementation, we adaptively select the reference poses in accordance with a desired nominal repeating velocity; however, now the reference poses, $\Trefk$, are selected directly from the teach path instead of the \ac{BIT*} solution. 

\subsection{Solution Method}

Given the optimization problem defined in \eqref{test}, our next step is to devise a method for effectively enforcing the series of box constraints related to the actuator limits and lateral corridor. To address this, we employ a technique known as the Barrier Method \cite{Anton2009}. 

Using the Barrier Method, we can capture the effects of constraints of the general form $\Matrix{x} \leq \Matrix{a}$ and $\Matrix{x} \geq \Matrix{a}$ as a series of logarithmic penalty terms. This transformation allows us to convert the constrained optimization problem into an unconstrained problem that can be solved efficiently. The squared logarithmic barrier function for an arbitrary inequality constraint is written as
\begin{subequations}
	\begin{align}
	&\Matrix{x} \leq \Matrix{a} \rightarrow
	\beta \sum_{i=1}^{\mathrm{dim}(\Matrix{x})} \ln^2 (\Matrix{a}_i - \Matrix{x}_i), \label{eq:4_10}\\
	&\Matrix{x} \geq \Matrix{a} \rightarrow
	\beta \sum_{i=1}^{\mathrm{dim}(\Matrix{x})} \ln^2 (\Matrix{x}_i - \Matrix{a}_i), \label{eq:4_11}
	\end{align}
\end{subequations}
%


%
\noindent where $\beta$ is a tunable scalar parameter to weight the influence of the barrier function. As the value of $\beta$ decreases, the barrier function approaches zero for all states $\Matrix{x}_i$ that satisfy the inequality constraint. However, the barrier function rapidly tends towards infinity as the state variables approach the constraint boundary. By adjusting the value of $\beta$, we can control the trade-off between satisfying the constraints and optimizing the objective function.

In practice, it is common to employ an iterative reweighting scheme to fine-tune the barrier parameters. Initially, a large value is assigned to the barrier parameters to heavily penalize violations of the constraints. The optimization problem is then solved iteratively, gradually reducing the barrier parameters in each iteration. This iterative process benefits from warm-starting, where the solution from the previous iteration is used as the initial guess for the current iteration. By carefully reducing the barrier parameters and avoiding abrupt changes, the optimization process converges to an optimal solution that balances constraint satisfaction and objective function optimization.

To apply \eqref{eq:4_10} and \eqref{eq:4_11} to the velocity and lateral path constraints, we define the vectors $\vel$ and $\lat$ to augment the original cost function as

\begin{align} \label{eq:4_12}
&\vel = \bbm \beta \ln (\Matrix{u}_{\mathrm{max},1} - \Matrix{u}_1) \\ \vdots \\
\beta \ln (\Matrix{u}_{\mathrm{max},2K} - \Matrix{u}_{2K}) \\
\beta \ln (\Matrix{u}_1 - \Matrix{u}_{\mathrm{min},1}) \\ \vdots \\
\beta \ln (\Matrix{u}_{2K} - \Matrix{u}_{\mathrm{min},2K}) \ebm, \\
&\lat = \bbm \beta \ln (\db_{1,r} - \y_1) \\ \vdots \\
\beta \ln (\db_{K,r} - \y_{K}) \\
\beta \ln (\y_1 + \db_{1,l}) \\ \vdots \\
\beta \ln (\y_{K} + \db_{K,l}) \ebm.
\end{align}
%
%
%
%
We define $\V$ and $\W$ as diagonal weighting matrices, so we may write the augmented version of the cost function \eqref{test} in the lifted form
\begin{subequations}
	\begin{align} \label{eq:4_14a}
	&\begin{aligned} [t]
	J(\T ,\ub, \y) =
	\sum_{k=1}^K &\ln (\Trefk \Tk^{-1})^{\vee^T}
	\Matrix{Q}_k
	\ln (\Trefk \Tk^{-1})^{\vee} \\
	&+ \ub_k^T \Matrix{R}_k \ub_k 
	+ \vel^T
	\, \V \, \vel\\  
	&+ \lat^T \, \W \, \lat \\
	\end{aligned} \\
	&\begin{aligned} [b]
	&\mathrm{s.t.} \\
	&\Tkone =
	\exp \Big( (\Matrix{P}^{T} \uk)^\wedge h \Big) \Tk, \quad k = 1, 2, \hdots , K. \label{eq:4_14b}
	\end{aligned}
	\end{align}
\end{subequations}

\noindent Problem \eqref{eq:4_14a} can be solved efficiently with a Gauss-Newton method by linearizing the problem at an operating point and applying small perturbations to the Lie Algebra to simplify terms\footnote{Please refer to the Appendix for a detailed solution method using Gauss Newton optimization.}. In practice, we solve the \ac{MPC} problem directly using the \ac{STEAM} engine \cite{Anderson2015}, an iterative Gauss-Newton-style optimization library aimed at solving batch nonlinear optimization problems involving both Lie Group and Continuous-time components. \ac{STEAM} is by no means the fastest way of solving these types of problems; however, it is convenient to work with problems modelled in this form and allows us to maintain suitable control rates in excess of 30 Hz.
\section{Offline Experiments}
\subsection{Performance Metrics}
In teach-and-repeat applications, it is critical that when avoiding obstacles we limit lateral path error to best exploit the prior knowledge on terrain assessment. It is also important to maintain a similar sensor viewing angle throughout the trajectory for localization purposes. With these characteristics in mind, we propose two metrics to compare the relative quality of the path solutions produced over the course of our experiments. We compute the Root Mean Square Error (RMSE) for both the translation and rotation (heading) components relative to the reference path over 15 m segments of obstacle interactions and average the result across all trials. In the context of this study, obstacle interactions refer to any event where a newly appeared obstacle (as labelled by terrain-change detection) causes the robot to deviate from the original reference path to avoid a collision.


The relevance of RMSE for translation error is straightforward, as deviations from the reference path during the repeat phase pose an increased risk. However, the use of RMSE to measure heading error requires further explanation. In visual teach-and-repeat systems, it is important to account not only for the robot’s position but also for the viewpoint angle, which is crucial for successful feature matching to the reference path. When the robot deviates from the path to avoid an obstacle, our goal is to minimize both the maximum heading error encountered and the overall average heading error required to navigate around the obstacle.
A non-zero average heading error is expected and necessary over the course of an obstacle interaction. However, with all else being equal, a trajectory that is on average able to maintain a more similar viewpoint angle to the reference path may be more desirable to mitigate risks associated with loss of localization. Thus the RMSE heading error can be a useful tool for attempting to benchmark motion planner trajectories against one another.


\subsection{Unit Testing of New Edge Cost}

We demonstrate the benefits of the lateral edge metric by testing BIT* on several representative obstacle-avoidance problems, both with and without the extensions proposed in Section III. For fair comparison, we implemented our own C++ version of BIT* in accordance with Gammell et al. \cite{Gammell2015}, using parameters of 150 samples per batch and an RGG constant of 1.1. Our extended implementation, Lateral BIT*, uses the laterally weighted edge-cost metric \eqref{eq:eight} with $\alpha = 0.5$  and the rectangular approximation of the informed sampling region. All experiments were conducted on a Lenovo Thinkpad laptop with 11th Gen Intel(R) Core(TM) i7-11800H @ 2.3GHz. For the standard BIT* implementation, we can force the use of a pure Euclidean distance edge metric by simply adjusting $\alpha=0$, and switch to using the ellipsoidal sampling region described by Gammell et al. \cite{Gammell2014}. 

\begin{figure*}[t]
	\addtolength{\abovecaptionskip}{-4pt}
	\centering
	\includegraphics[scale=0.788]{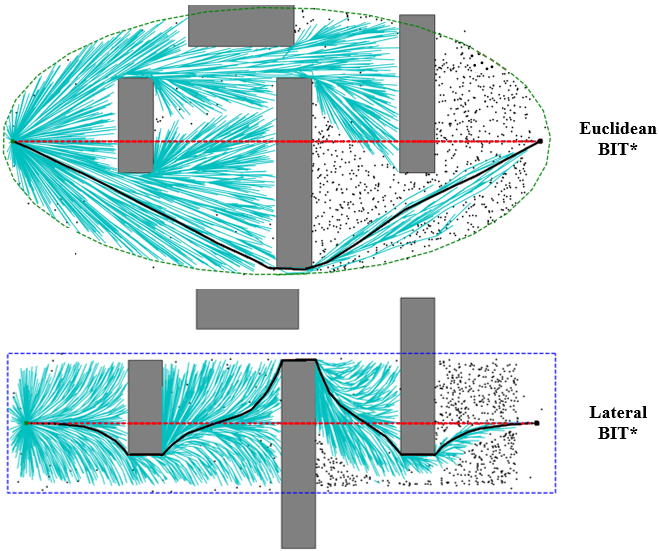}
	\caption{A comparison of the planning trees generated running BIT* with the Euclidean edge metric (top) and with the laterally weighted metric (bottom) in a representative test environment. Our edge metric encourages the final plan, shown in black, to avoid obstacles while remaining close to the red reference path on the $p$-axis.}
	\label{figure5}
	\vspace{-1em}
\end{figure*}

To study the influence of the new edge metric in isolation, we posed a series of ten simulated planning problems where the reference path is composed of a horizontal line 15 m in length on the $x$-axis, making the curvilinear and Euclidean path representations identical. We allow both versions of the algorithm to find converged path solutions connecting the starting pose to the goal and evaluate the solution both quantitatively and qualitatively. A representative example of the output paths and associated BIT* trees on one test problem is shown in Fig. \ref{figure5}.

In analyzing Fig. \ref{figure5}, we see some important differences in the exploration strategies of the two methods. Using our lateral edge-cost metric, BIT* tends to naturally generate smoothly curving solutions, spending the most time exploring paths near to the reference while balancing forward progress with reducing lateral error. In contrast, the standard BIT* algorithm settles on the direct shortest-distance solution. While efficient, in practice this path could be a higher-risk manoeuvre due to the additional localization and terrain-assessment uncertainty incurred when away from the reference path. We calculate the RMSE metrics for both implementations and summarize the results in Table 1.

\begin{table}[!b]
	\captionsetup{skip=0pt} 
	\caption{Path Planner Error Analysis For Simulation Experiments}
	\vspace{-6pt}
	\label{BIT_Simulated_RMSE}
	\begin{center}
		\scalebox{0.64}{
			\begin{tabular}{|c||c|c||c|c|}
				\hline
				\textbf{Exp. \#}	& \specialcell[c]{\textbf{Lateral}\\\textbf{RMSE [cm]}\\\textbf{(Lateral BIT*)}} & \specialcell[c]{\textbf{Lateral}\\\textbf{RMSE [cm]}\\\textbf{(Euclidean BIT*)}} & \specialcell[c]{\textbf{Heading}\\\textbf{RMSE [deg]}\\\textbf{(Lateral BIT*)}} & \specialcell[c]{\textbf{Heading}\\\textbf{RMSE [deg]}\\\textbf{(Euclidean BIT*)}}\\
				\hline
				1. & 8.52 & 26.80 & 29.46 & 35.71\\
				\hline
				2. & 10.11 & 22.18 & 32.95 & 29.61\\
				\hline
				3. & 9.75 & 24.73 & 30.25 & 34.74\\
				\hline
				4. & 9.90 & 31.03 & 25.02 & 31.87\\
				\hline
				5. & 7.16 & 22.61 & 23.80 & 25.39\\
				\hline
				6. & 11.24 & 25.10 & 41.31 & 32.73\\
				\hline
				7. & 10.81 & 26.79 & 29.11 & 35.42\\
				\hline
				8. & 9.51 & 24.96 & 30.70 & 31.12\\
				\hline
				9. & 8.90 & 27.03 & 27.74 & 43.67\\
				\hline
				10. & 12.37 & 23.72 & 35.78 & 28.90\\
				\hline
				\textbf{Mean:} & \textbf{9.83} & \textbf{25.50} & \textbf{30.61} & \textbf{32.92}\\
				\hline
				
		\end{tabular}}
	\end{center}
\end{table}

As we would expect, the use of our laterally weighted edge metric considerably reduces the average lateral error with respect to the reference path from 0.254 m to 0.098 m. We also see a small improvement on the heading error, likely due to the fact the lateral planner tends to spend more time exactly following the reference path. 

\begin{figure}[t]
	\centering
	\addtolength{\abovecaptionskip}{-4pt}
	\includegraphics[scale=1.22]{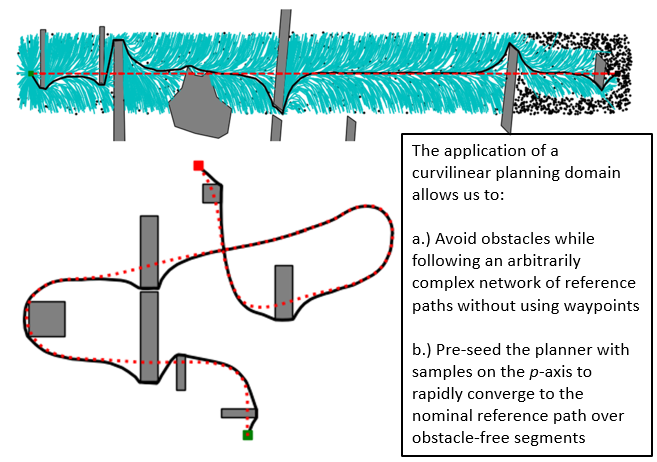}
	\caption{Lateral BIT* planning in a curvilinear representation of the space (top) to find a global obstacle avoidance path solution in Euclidean space (bottom) along a complex reference path with many obstacles. The use of the curvilinear configuration space offers two main advantages: no intermediate waypoints are needed to follow the initial reference path, and we can always predict the optimal solution in the absence of obstacles (a straight horizontal line). This allows us to preseed the path with samples, leading to faster planner convergence.}
	\label{figure6}
	\vspace{-1em}
\end{figure}

Using straight-line reference paths, it is easy to see that the lateral edge cost encourages solutions with desirable path properties. However, we can further exploit the curvilinear coordinate space to produce similarly smooth plans on more intricate reference paths. In Fig. \ref{figure6}, we initialize the planner on a complex path taken from real teach data that includes a variety of sharp curves, a path crossing, and several difficult obstacles. Despite the challenges, our planner is able to converge to a desirable solution using only a single goal and no intermediate waypoints. As Euclidean BIT* tends to generate solutions that traverse singularity regions in curvilinear space (this subsequently produces discontinuous Euclidean paths), we are not able to directly provide a comparison to this result.

\begin{figure}[t]
	\centering
	\addtolength{\abovecaptionskip}{-4pt}
	\includegraphics[scale=0.5]{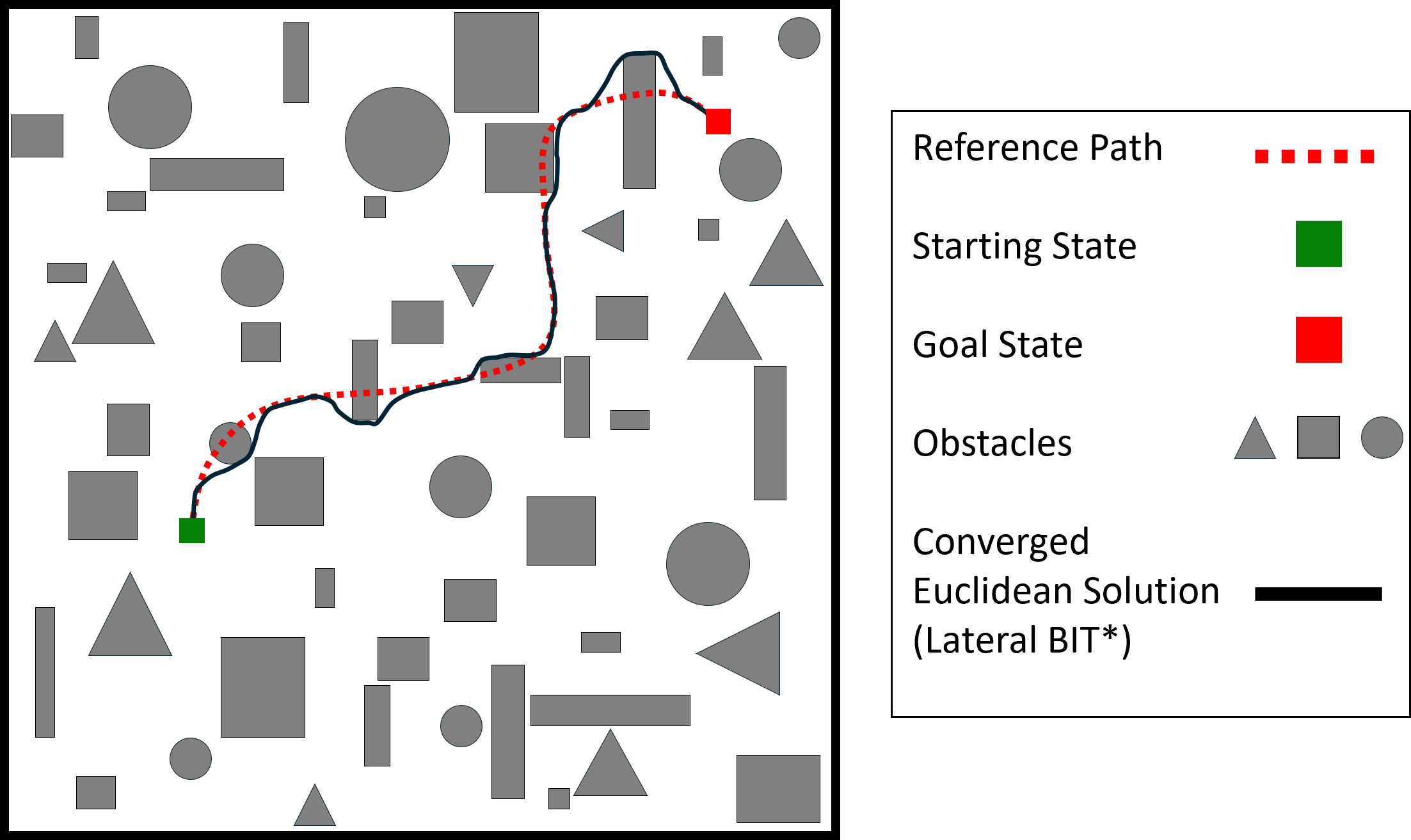}
	\caption{Lateral BIT* planning in a representative randomly generated square simulated environment of 25 m x 25 m scattered with 50 obstacles (grey). A smooth initial reference path connecting the starting square (green) and the goal (red) is created using cubic splines. Lateral BIT* then uses this spline as the teach path and attempts to follow the path while avoiding any new obstacles and minimizing lateral deviations.}
	\label{fig:100Sim_Ex}
	\vspace{-1em}
\end{figure}

To expand our offline simulation results to a broader range of reference paths, we conducted 100 additional planning experiments under the following conditions. A start and goal point, randomly placed at least 15 meters apart, were set in a square environment with 25-meter sides. A random reference path was generated using cubic splines to emulate the curvilinear space. Fifty obstacles of varying sizes and shapes were then randomly distributed throughout the environment, and our laterally weighted BIT* planner was run to convergence. An example trial is shown in Fig. \ref{fig:100Sim_Ex}.

Unfortunately, as there is not a unique map from Euclidean space to curvilinear space, we are unable to easily visualize obstacles and paths in the curvilinear space. Instead we opt to focus on quality of the final Euclidean result. Of the 100 trials, 87 successfully reached the goal, achieving an average lateral RMS error of 11.02 cm and an average heading RMSE of 26.2 degrees, comparable to previous straight-line reference trials.  The 13 failures occurred when the planner returned early that no solution could be found without deviating more than the constant 2.5 m maximum lateral path deviation that was set for the trials. This was to be expected and represents a case where the significant changes in the environment presented an unsafe navigation situation that we opt to stop and flag for operator intervention to prevent localization failures and collisions in unknown terrain.

\subsection{Runtime Analysis}

\begin{figure}[b]
	\centering
	\includegraphics[scale=0.92]{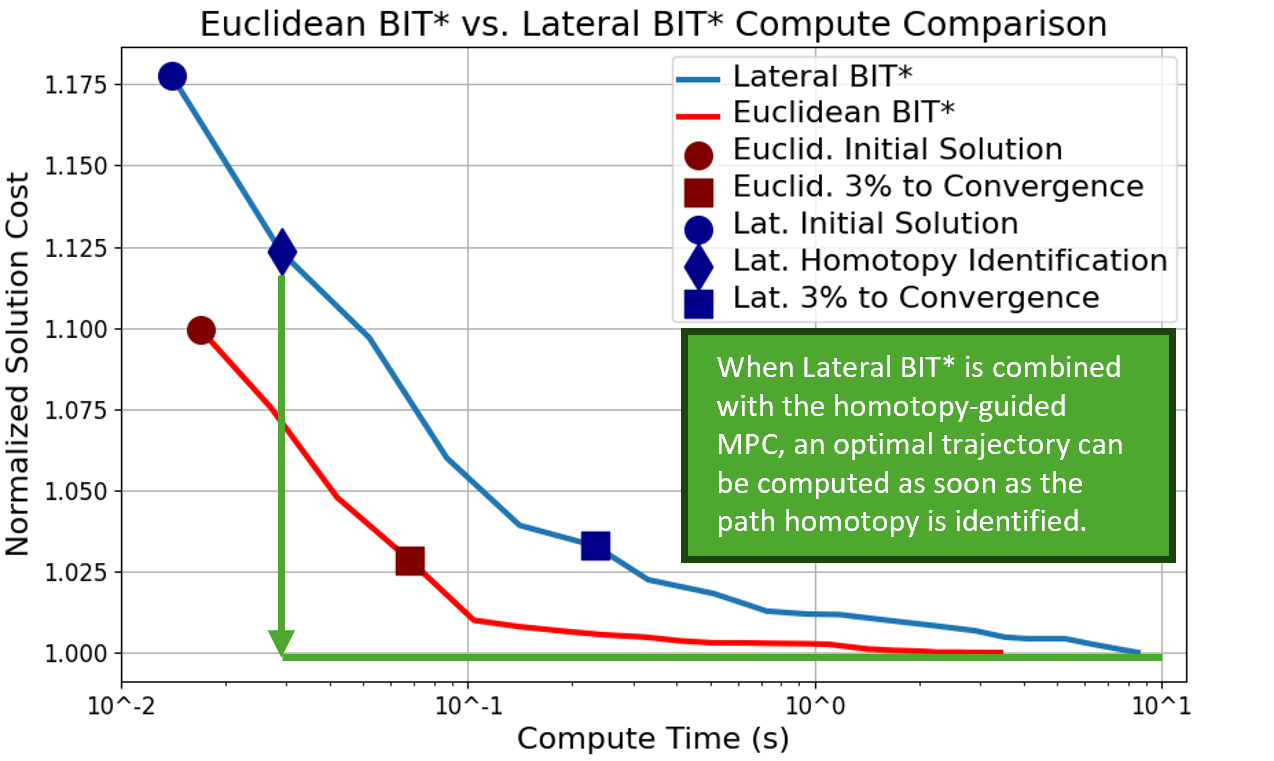}
	\caption[Euclidean BIT* vs. Lateral BIT* Compute Comparison]{In this figure we compare the compute characteristics of the Euclidean BIT* planner to the Lateral BIT* planner. We find that generally the Euclidean edge-cost metric converges faster with marginally higher early solution quality. However, it is important to recognize that for the key compute metrics that most heavily influence the performance our architecture, the initial solution time and the homotopy class identification time, Lateral BIT* performs comparably. After Lateral BIT* identifies the correct homotopy class (blue diamond), our homotopy-guided mpc can immediately be used to generate an optimal trajectory and jump to an effective normalized solution cost of 1.000.}
	\label{BIT_Compute}
\end{figure}

Regardless of path-following capability, real-time performance is crucial for the path-planning module. In Fig. \ref{BIT_Compute}, we analyze the runtime performance and observe that our extended version of \ac{BIT*} generally generates initial solutions and converges to an optimal result at a slower pace compared to the Euclidean version of \ac{BIT*}. This outcome was anticipated based on the qualitative path observations in Fig. \ref{figure5}, where a higher number of samples per batch is required to generate the observed `weaving' plans. Intuitively, a high sample density is necessary to refine the smooth curves that are characteristic of our extensions, which leads to slower convergence.

However, it is important to highlight that the differences in compute times are relatively small. Despite this, our Lateral \ac{BIT*} implementation still manages to find initial solutions to challenging planning problems in just 17 ms. Furthermore, it achieves convergence to within 3\% of the optimal solution cost in just 233 ms on average, ensuring the generation of smooth output paths in a timely manner.

The true strength of our planner lies in its average time for correct homotopy-class identification. This metric effectively measures the time taken by the planner to generate a solution that avoids obstacles and can be continuously deformed into the optimal solution without colliding with obstacles. On average, our planner is able to identify these correct homotopy solutions sufficient for running homotopy-guided MPC in just 23 ms, which is nearly three times faster than the Euclidean planner can approach a converged solution suitable for direct-tracking MPC.

\section{Online Experiments}

\subsection{Introduction}
Our final set of experiments aims to assess the effectiveness of our entire system in real-world obstacle-avoidance scenarios. We integrated our obstacle-avoidance architecture into the \ac{VTR3} codebase and conducted extensive autonomy tests in two diverse environments as shown in Fig. \ref{Test_Environments}. These environments presented contrasting challenges: the first involved navigating a relatively flat prairie terrain with a simple single loop, while the second featured a more complex path in a valley with varying elevations. 

\begin{figure*}[t]
	\centering
	\includegraphics[scale=1.1]{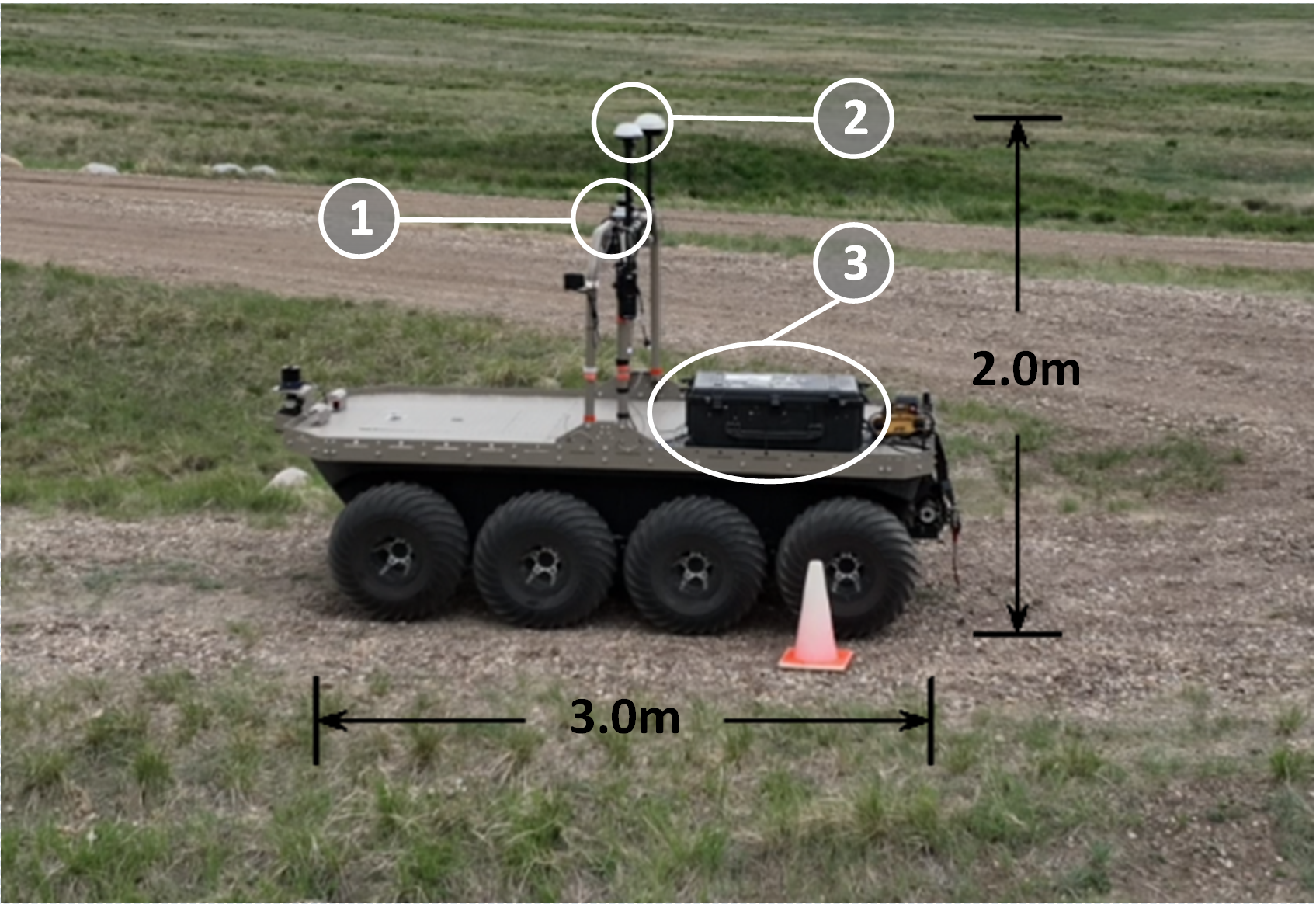}
	\caption[Robot Setup]{The ARGO Atlas J8 Unmanned ground vehicle used to conduct all field evaluation: (1) Ouster OS1 LiDAR 1024x64 @ 10Hz, (2) Hemisphere Dual Differential GPS, (3) Internal Lenovo Thinkpad laptop with 11th Gen Intel(R) Core(TM) i7-11800H @ 2.3GHz.}
	\label{ARGO_Test_Setup}
	\vspace{-1em}
\end{figure*}

In each case, the robot was initially manually driven to teach an obstacle-free network of paths. Subsequently, various obstacles were introduced throughout the path to obstruct the predefined routes. Through a series of experiments, we repeated the taught paths and analyzed the resulting trajectories produced by our two motion-planning architectures. The experiments were performed in collaboration with \ac{DRDC}, utilizing an ARGO Atlas J8 electric differential-drive robot, Fig.\ref{ARGO_Test_Setup}, under identical configurations to those used for the offline experiments.

\begin{figure}[t]
	\centering
	\captionsetup[subfloat]{labelfont=scriptsize,textfont=scriptsize}
	\subfloat[Easy Scenario]{\includegraphics[scale=0.555]{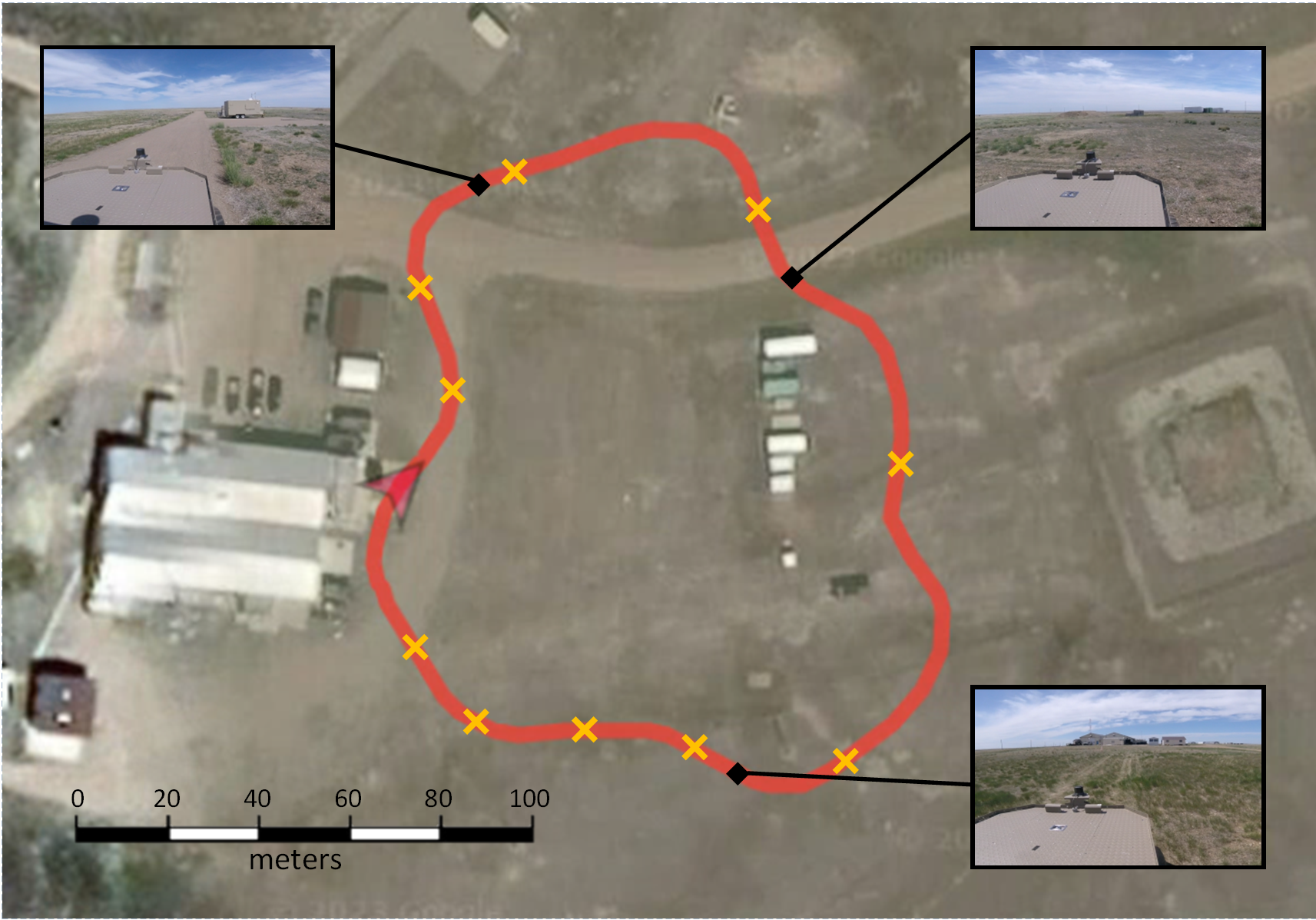}}
	\hfill
	\subfloat[Hard Scenario]{\includegraphics[scale=0.585]{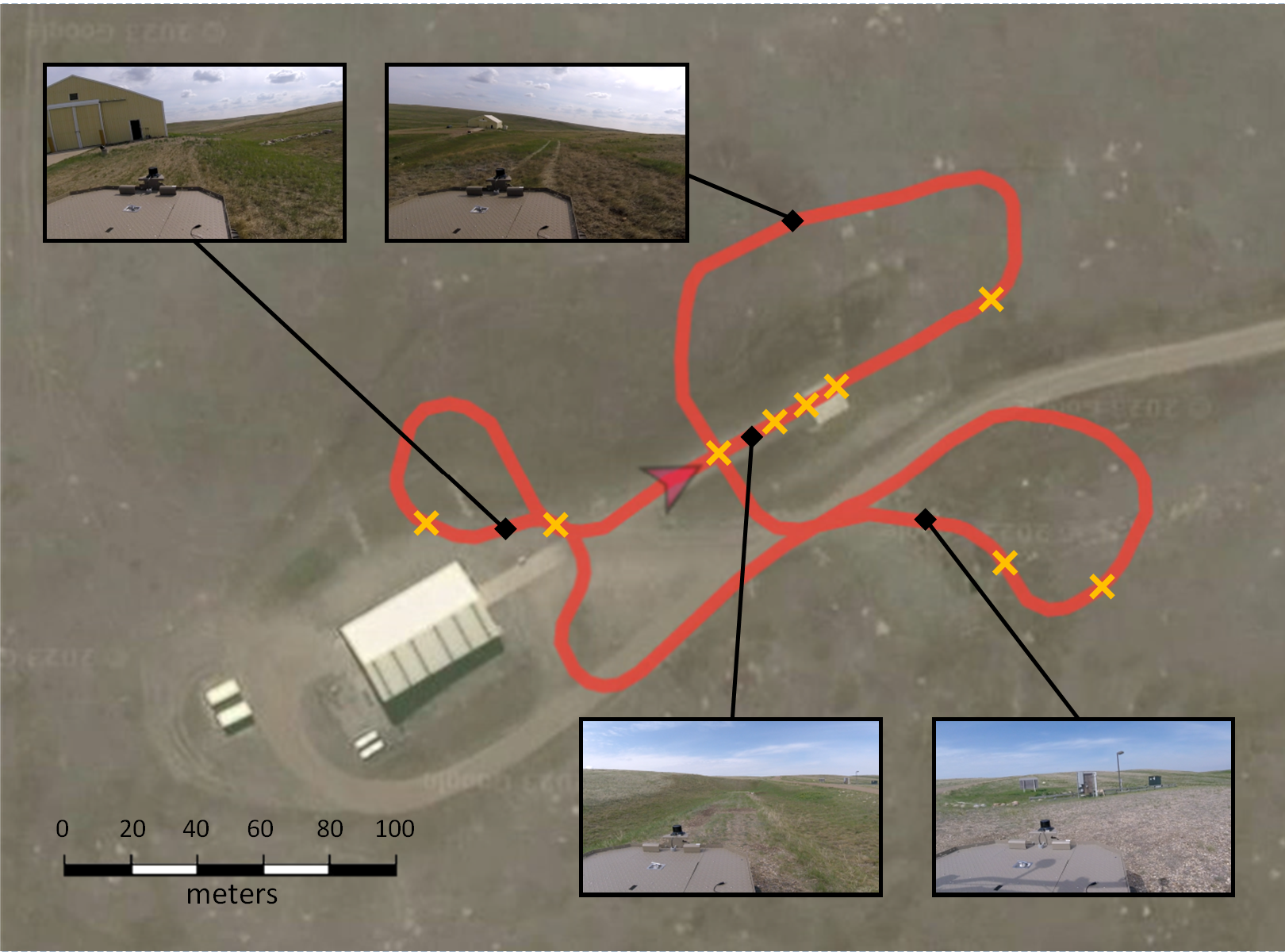}}
	
	\caption[Easy Obstacle Avoidance Scenario Test Site]{We evaluate our local obstacle-avoidance system at the Experimental Proving Ground on Canadian Forces Base Suffield. In the first scenario, (a), we taught a short loop on reasonably flat and open terrain. In the second scenario, (b), we taught a long and intricate route in the middle of a valley consisting of many elevation changes and diverse terrain. We then introduced a series of obstacles (yellow crosses) to the reference paths, forcing the robot to deviate from the teach path to complete subsequent repeat objectives. A demo video from the trials is available at \url{https://www.youtube.com/watch?v=wudAHTGKWyQ}.}
	\label{Test_Environments}
	\vspace{-1em}
\end{figure}


\subsection{Evaluation Strategy and Metrics}

In our evaluations, we emphasize the importance of obstacle-avoidance trajectories that strike a balance between minimizing lateral deviations from the taught path and ensuring smooth forward progress. This balance is crucial due to the prior knowledge we have about the taught path's collision-free nature, as verified by a human operator. By closely following the reference path, we maximize our chances of avoiding unforeseen obstacles and improve localization performance in \ac{VTR3} despite perception uncertainties.

To assess the performance, we report the \ac{RMS} lateral and heading errors obtained from both the differential \ac{GPS}\footnote{Despite recording GPS as an alternative source of position estimate for error calculation, we do not use the GPS position in any component of our system and generally assume we are operating in GPS-denied environments.} and the \ac{VTR3} state estimation system during the repeat experiments. However, it is important to note that these values become subjective, as they are influenced by the size and density of obstacles encountered during each repeat. To address this, we propose a new metric called the `obstacle interaction' \ac{RMS} lateral and heading error that focuses on a specific obstacle's influence on path deviation. We define an obstacle interaction as the length of the path extending 5 m on either side of the obstacle. By comparing the obstacle interaction RMSE with the error in obstacle-free sections, we gain insights into the relative amount of path deviation the planner uses to successfully avoid the obstacle.

Additionally, we consider the average maximum lateral deviation per obstacle interaction. This measure, when evaluated in the context of the lateral extent of the obstacle obstructing the reference path, provides an indication of how effectively the planner avoids obstacles while staying close to the reference path. We are particularly interested in observing the consistency of the maximum lateral deviation across obstacles with varying geometries.

When comparing the two motion planners, the direct-tracking \ac{MPC} and the homotopy-guided \ac{MPC}, we propose using the average robot path curvature as a metric to evaluate the relative smoothness of these approaches. Our rationale is that if both motion planners can maintain similar desirable lateral error characteristics during obstacle avoidance, the path with the lowest average curvature is likely to be preferable in terms of energy efficiency.

Finally, one of our most critical and intuitive metrics of interest is the obstacle-avoidance rate. This rate is calculated by dividing the number of successful obstacle interactions (those that are collision-free) by the total number of obstacles encountered on the route. It serves as a reliable indicator of the improvement in the long-term autonomy of the \ac{VTR3} system, as our goal is to minimize the need for operator interventions during autonomous navigation.

\subsection{Results (Easy Scenario)}

\begin{figure}[b]
	\centering
	\includegraphics[scale=0.71]{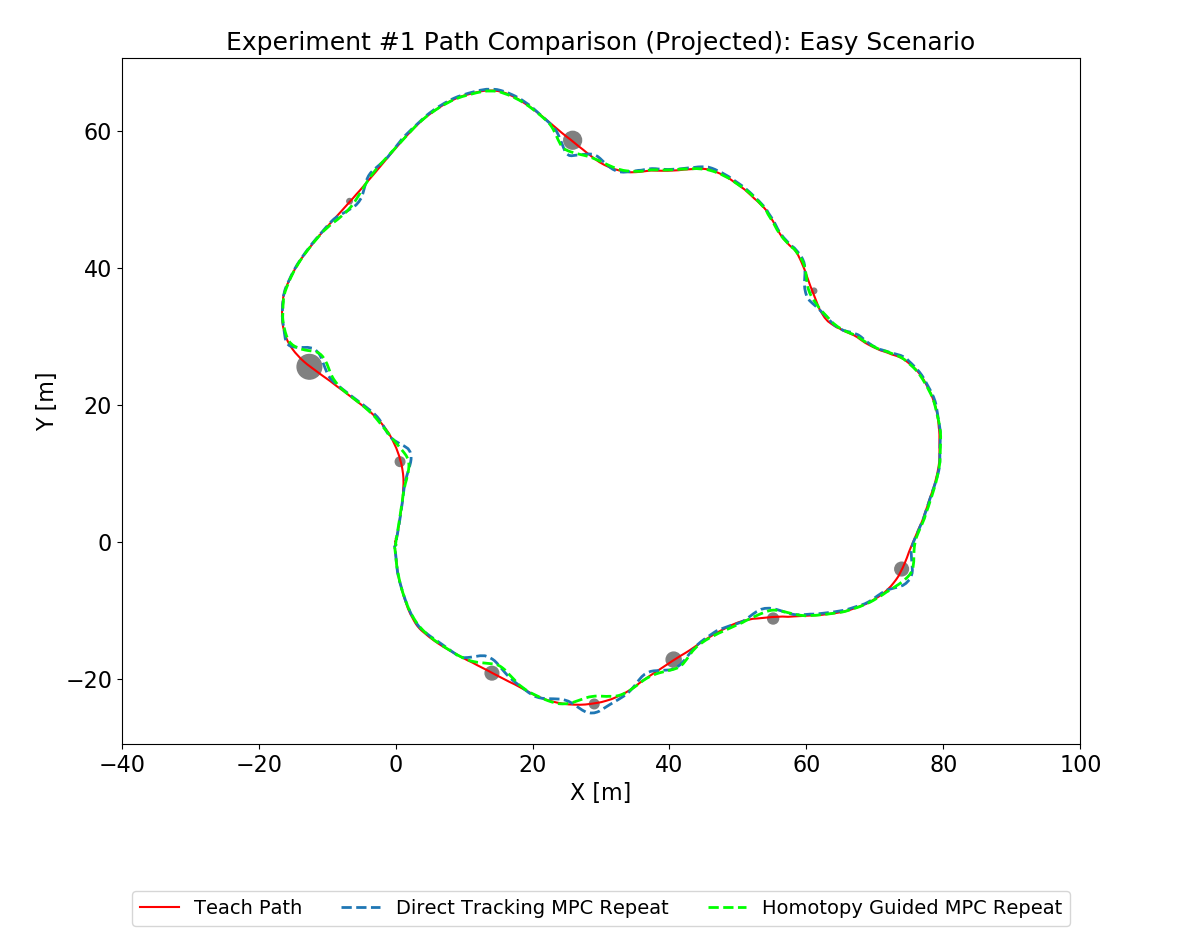}
	\caption[Easy Obstacle Avoidance Scenario Path Comparisons]{A plot of the repeat trajectories overlaid with the teach path as reported by \ac{GPS} ground truth. We see that the homotopy-guided \ac{MPC} solution (green), generally avoids obstacles with more consistent obstacle interaction profiles and with less overall lateral path deviation then the direct-tracking approach (blue).}
	\label{B15_Obs_Repeat_Path}
	\vspace{-1em}
\end{figure}

\begin{figure}[t]
	\centering
	\captionsetup[subfloat]{labelfont=scriptsize,textfont=scriptsize}
	\subfloat[Small trailer]{\includegraphics[height=7.0cm]{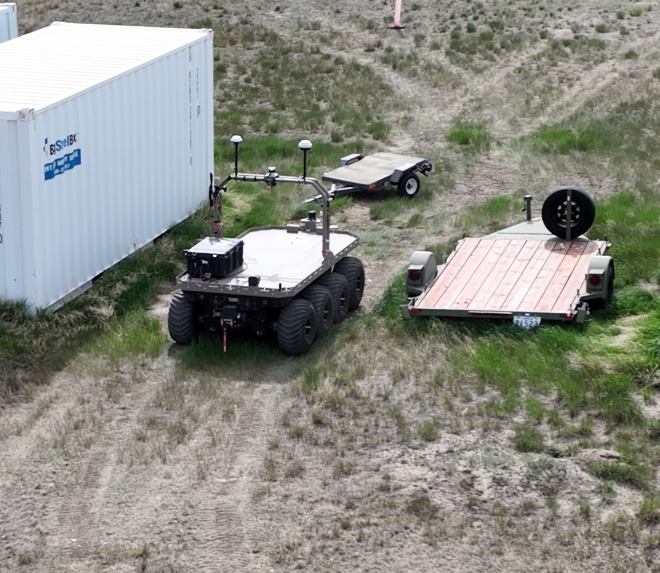}\label{fig:B15_Obs_Interaction_Ex1}}
	\hfill
	\subfloat[Wooden pallet]{\includegraphics[height=7.0cm]{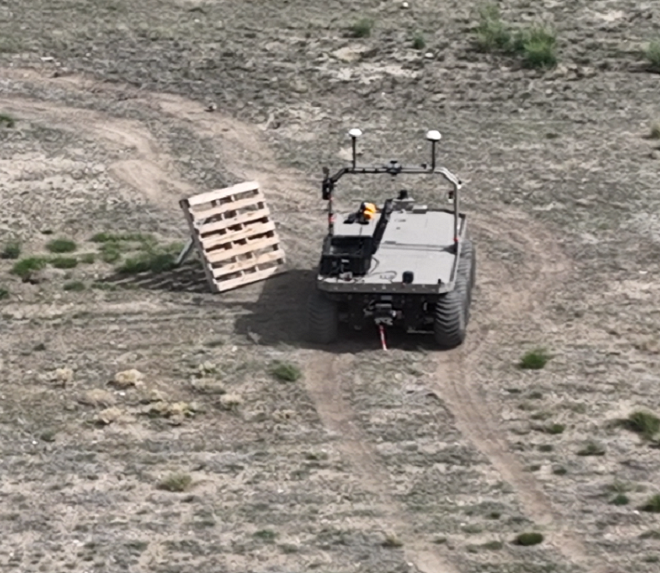}\label{fig:B15_Obs_Interaction_Ex3}}
	
	\caption[Easy Obstacle Avoidance Scenario Examples]{Some typical obstacle interactions encountered along the reference path in the Easy Scenario.}
	\label{B15_Obs_Interaction_Ex}
	\vspace{-1em}
\end{figure}

Upon repeating the loop both without obstacles and with obstacles, we present the actual robot path relative to the taught reference path (as reported by \ac{GPS}) across the loop in Fig. \ref{B15_Obs_Repeat_Path}. The locations and sizes of the obstacles have been marked for reference, and some examples of obstacle interactions are provided in Fig. \ref{B15_Obs_Interaction_Ex}. In both the direct-tracking \ac{MPC} and the homotopy-guided \ac{MPC} implementations, the robot successfully completed the loop 5 times at a target reference speed of 1.25 m/s, achieving a 100\% obstacle-avoidance rate. Throughout this experiment, the robot covered a cumulative distance of 1.5 km, navigating through 50 obstacles without any collisions or operator interventions. While both controllers accomplished our primary goal in this less-challenging scenario, we will now delve into the relative differences and trajectory characteristics of the two methods.

\begin{figure}[t]
	\centering
	\captionsetup[subfloat]{labelfont=scriptsize,textfont=scriptsize}
	\subfloat[Direct-Tracking MPC]{\includegraphics[height=6.59cm]{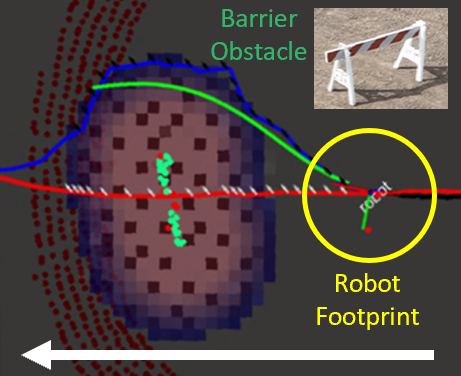}\label{fig:B15_Obs_Interaction_Ex_RViz2}}
	\hfill
	\subfloat[Homotopy-Guided MPC]{\includegraphics[height=6.59cm]{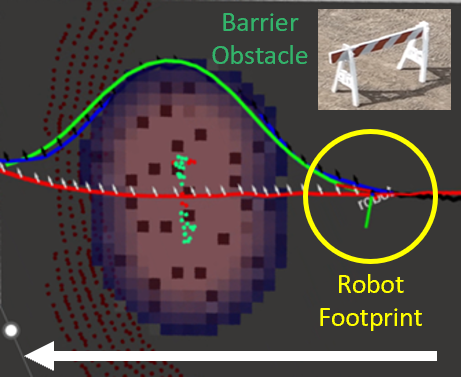}\label{fig:B15_Obs_Interaction_Ex_RViz3}}
	
	\caption[Easy Obstacle Avoidance Scenario Examples]{
		In this figure, we compare the solutions of the two controller architectures over a representative obstacle-avoidance scenario with a barrier. In both cases, the current planner solution (blue path) provides an initial path leading the robot around the obstacle and back to the reference path (red path). In (a), the direct-tracking MPC produces a smooth trajectory (green path), to track the BIT* solution, however, due to infeasible kinematics in the planner solution, there is some tracking error that takes the robot dangerously close to the obstacle. In contrast, the homotopy-guided MPC (b), generates a tight and safe trajectory around the obstacle using a similar planner solution (green path).}
	\label{B15_Obs_Interaction_Ex_RViz}
	\vspace{-1em}
\end{figure}

Qualitatively, we can compare the set of trajectories produced over one obstacle interaction in Fig. \ref{B15_Obs_Interaction_Ex_RViz}. In both cases, the current planner solution (blue path) leads the robot around the obstacle and back to the reference path (red path). The direct-tracking MPC, \eqref{fig:B15_Obs_Interaction_Ex_RViz2}, generates a smooth trajectory (green path) to track the BIT* solution, though due to kinematic infeasibilities in the planner's solution, there is a tracking error that brings the robot dangerously close to the obstacle. In contrast, the homotopy-guided MPC, \eqref{fig:B15_Obs_Interaction_Ex_RViz3}, produces a tighter, safer trajectory around the obstacle, demonstrating better obstacle avoidance with a similar planner solution.

To assess the performance of the planners during obstacle interactions, we calculated the \ac{RMS} lateral and heading errors for obstacle-free repeats of the environment, serving as a baseline for path-tracking performance. For this particular path network, the direct-tracking \ac{MPC} achieved a lateral \ac{RMS} error of 2.50 cm, while the homotopy-guided \ac{MPC} yielded a slightly lower error of 2.21 cm. Similarly, the baseline heading \ac{RMS} errors were 4.5 deg and 4.0 deg, respectively.

\begin{figure*}[t]
	\centering
	\includegraphics[scale=0.94]{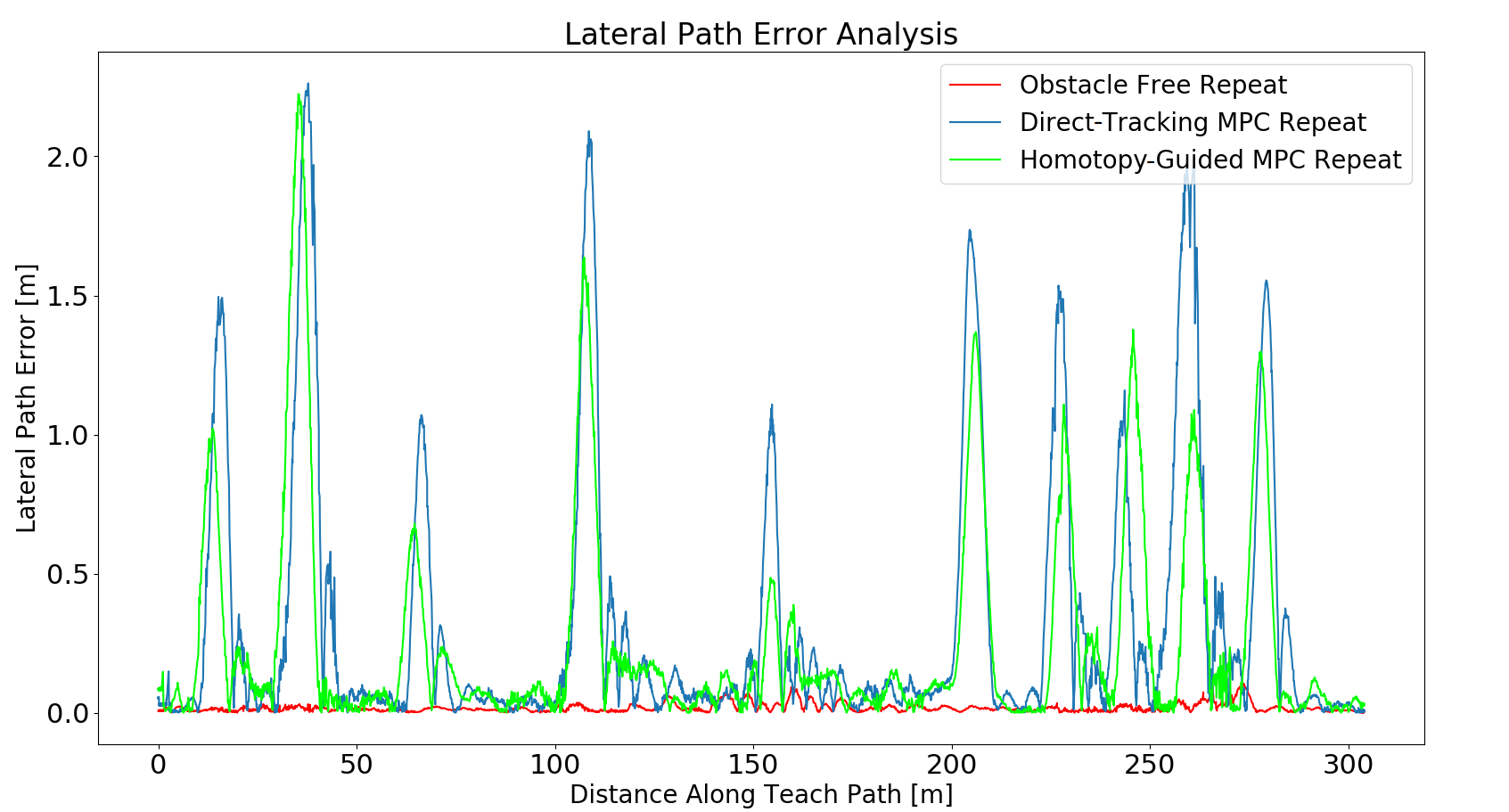}
	\caption[Easy Obstacle Avoidance Scenario Lateral Error Analysis]{A comparison of the lateral path errors across the obstacle-avoidance repeats for the two proposed control architectures on the easy scenario.}
	\label{B15_Obs_Lateral_Error_Plots}
\end{figure*}

\begin{figure*}[t]
	\centering
	\includegraphics[scale=1.78]{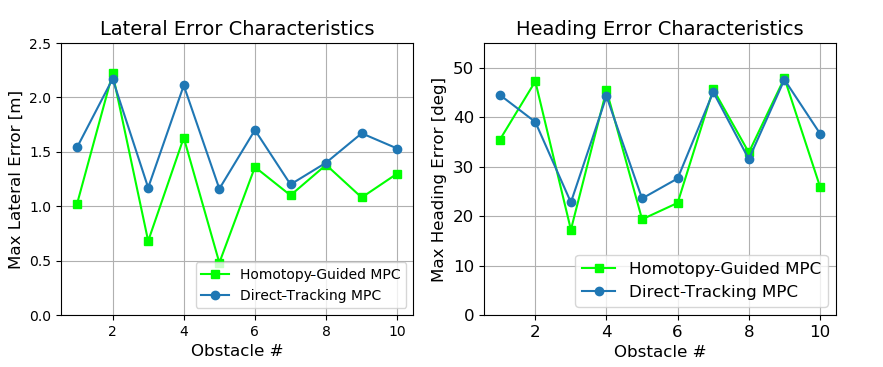}
	\caption[Easy Obstacle Avoidance Scenario Error Profile]{The maximum tracking error characteristics on an individual obstacle interaction basis on the easy scenario.}
	\label{B15_Obs_Max_Error_Plots}
\end{figure*}

Next, we recalculated the \ac{RMS} error metrics specifically for the obstacle-avoidance interactions, focusing on the 10 m segments of the path centered around each obstacle. The results of the obstacle interaction analysis for one loop repetition of each controller are presented in Table \ref{B15_Easy_Obs_Interaction_Table}. We provide the extent of the obstacle on the path for reference as this directly influences how far the robot must deviate from the path during an obstacle interaction and depends on both the size of the robot and obstacle. Our findings indicate that both controllers exhibit comparable performance in this aspect, with the homotopy-guided \ac{MPC} demonstrating a reduction in obstacle interaction errors compared to the direct-tracking \ac{MPC}, as anticipated. The obtained data aligns with our qualitative observations, affirming that the planner effectively avoids significant translation and rotation errors relative to the extent of the encountered obstacles, while avoiding any adverse effects on localization. 

\begin{table}[b]
	\vspace{-6pt}
	\caption{Obstacle Avoidance Error Table: Easy Scenario}
	\vspace{-6pt}
	\label{B15_Easy_Obs_Interaction_Table}
	\begin{center}
		\scalebox{0.75}{
			\begin{tabular}{|c|c||c|c||c|c|}
				\hline
				\multicolumn{2}{|c||}{} & \multicolumn{2}{c||}{\textbf{Direct-Tracking MPC}} & \multicolumn{2}{c|}{\textbf{Homotopy-Guided MPC}} \\
				\cline{3-6}
				\multicolumn{2}{|c||}{\specialcell[c]{\textbf{Obstacle \#}\\\textbf{Extent [m]}\\}} &
				\specialcell[c]{\textbf{Lateral}\\\textbf{RMSE [m]}} & \specialcell[c]{\textbf{Heading}\\\textbf{RMSE [deg]}} &
				\specialcell[c]{\textbf{Lateral}\\\textbf{RMSE [m]}} & \specialcell[c]{\textbf{Heading}\\\textbf{RMSE [deg]}} \\
				\hline
				1. & 0.7 & 1.54 & 22.98 & 0.62 & 14.37\\
				\hline
				2. & 1.8 & 2.17 & 34.69 & 1.46 & 27.38\\
				\hline
				3. & 0.3 & 1.17 & 17.63 & 0.42 & 8.81\\
				\hline
				4. & 1.3 & 2.11 & 21.85 & 1.02 & 21.74\\
				\hline
				5. & 0.2 & 1.16 & 16.47 & 0.27 & 10.50\\
				\hline
				6. & 1.0 & 1.70 & 15.39 & 0.87 & 14.87\\
				\hline
				7. & 0.8 & 1.20 & 18.52 & 0.61 & 17.16\\
				\hline
				8. & 1.1 & 1.40 & 18.87 & 0.81 & 18.12\\
				\hline
				9. & 0.7 & 1.67 & 20.94 & 0.66 & 17.40\\
				\hline
				10. & 1.0 & 1.53 & 21.23 & 0.84 & 14.94\\
				\hline
		\end{tabular}}
	\end{center}
	\vspace*{-0.5cm}
\end{table}

Additionally, we consider the maximum lateral deviation per obstacle interaction as an informative metric. By examining the peaks of the lateral error magnitude plot along the path for both planners in Fig. \ref{B15_Obs_Lateral_Error_Plots}, we find that, on average, the direct-tracking \ac{MPC} deviates laterally from the reference path by a maximum of $r$ + 0.675 m with a standard deviation of 0.254 m, while the homotopy-guided \ac{MPC} results in $r$ + 0.335 m with standard deviation 0.048 m. In this context, $r$ represents the lateral extent of the obstacle obstructing the reference path. Theoretically, the lower bound of this value should be exactly $r$ for collision-free avoidance, but knowing that our perception is not infallible and to ensure a small safety buffer, we introduce an obstacle-inflation tuning parameter of 0.3 m, resulting in a slight expected offset. Notably, the homotopy-guided motion planner is able to achieve maximum lateral deviation characteristics close to this bound and the small uncertainty in the measure is evident of consistent obstacle-avoidance behaviour across obstacles of varying sizes and geometries. In contrast, the direct-tracking \ac{MPC} tends to take a more conservative approach to avoiding obstacles, with less repeatability. This result can be visualized more clearly by plotting the maximum values during each obstacle interaction as in Fig. \ref{B15_Obs_Max_Error_Plots}.

Another aspect of comparison between the two motion planners is the average curvature of the robot trajectory across the repeats. Intuitively, the presence of obstacles leads to an increase in average path curvature, as the robot maneuvers to avoid collisions. As a baseline, we see that the obstacle-free average curvature of 0.093 m$^{-1}$ increases to 0.112 m$^{-1}$ for the direct-tracking \ac{MPC} and 0.105 m$^{-1}$ for the homotopy-guided \ac{MPC}, representing a 20.4\% and 12.9\% increase, respectively. 

\subsection{Results (Hard Scenario)}

\begin{figure*}[t]
	\centering
	\includegraphics[scale=0.89]{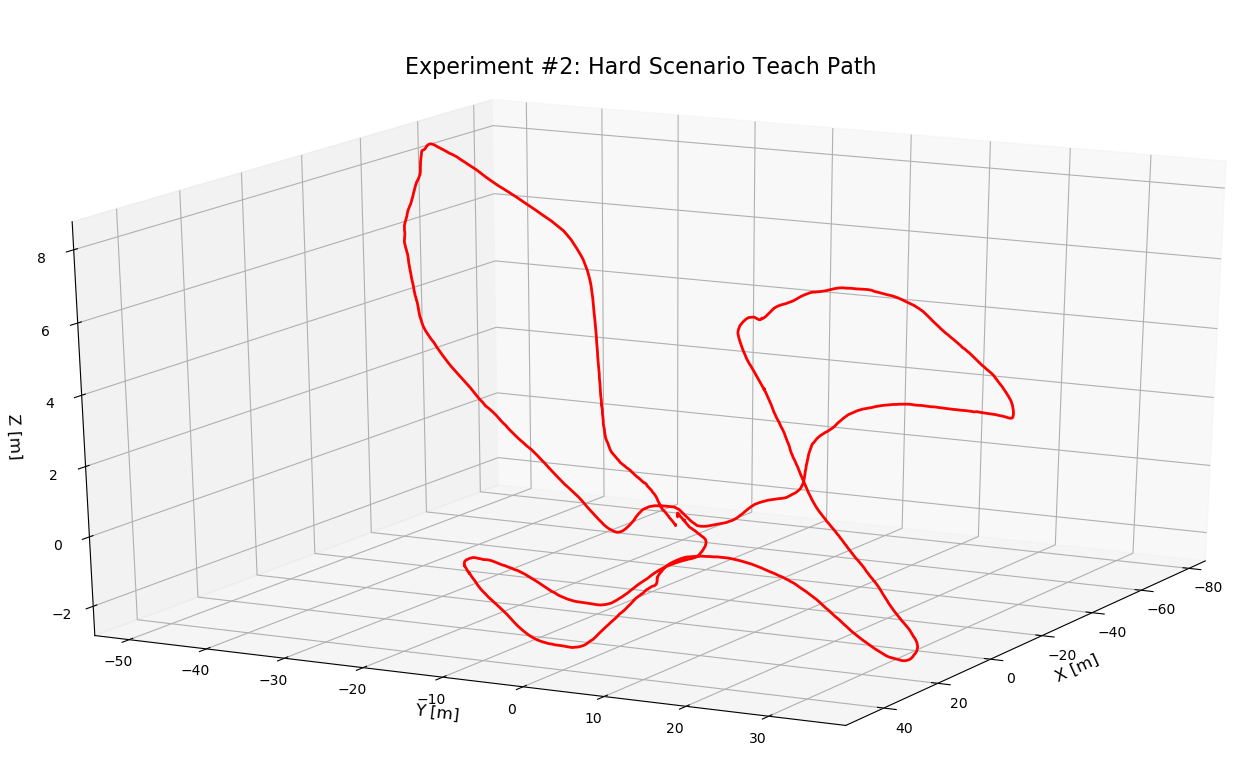}
	\caption[Hard Obstacle Avoidance 3D Reference Path]{The taught reference path showcasing the large elevation changes across the ``Hard Scenario" loop as reported by \ac{GPS}.}
	\label{OATS_Obs_Repeat_Path}
\end{figure*} 

\begin{figure*}[t]
	\centering
	\includegraphics[scale=0.83]{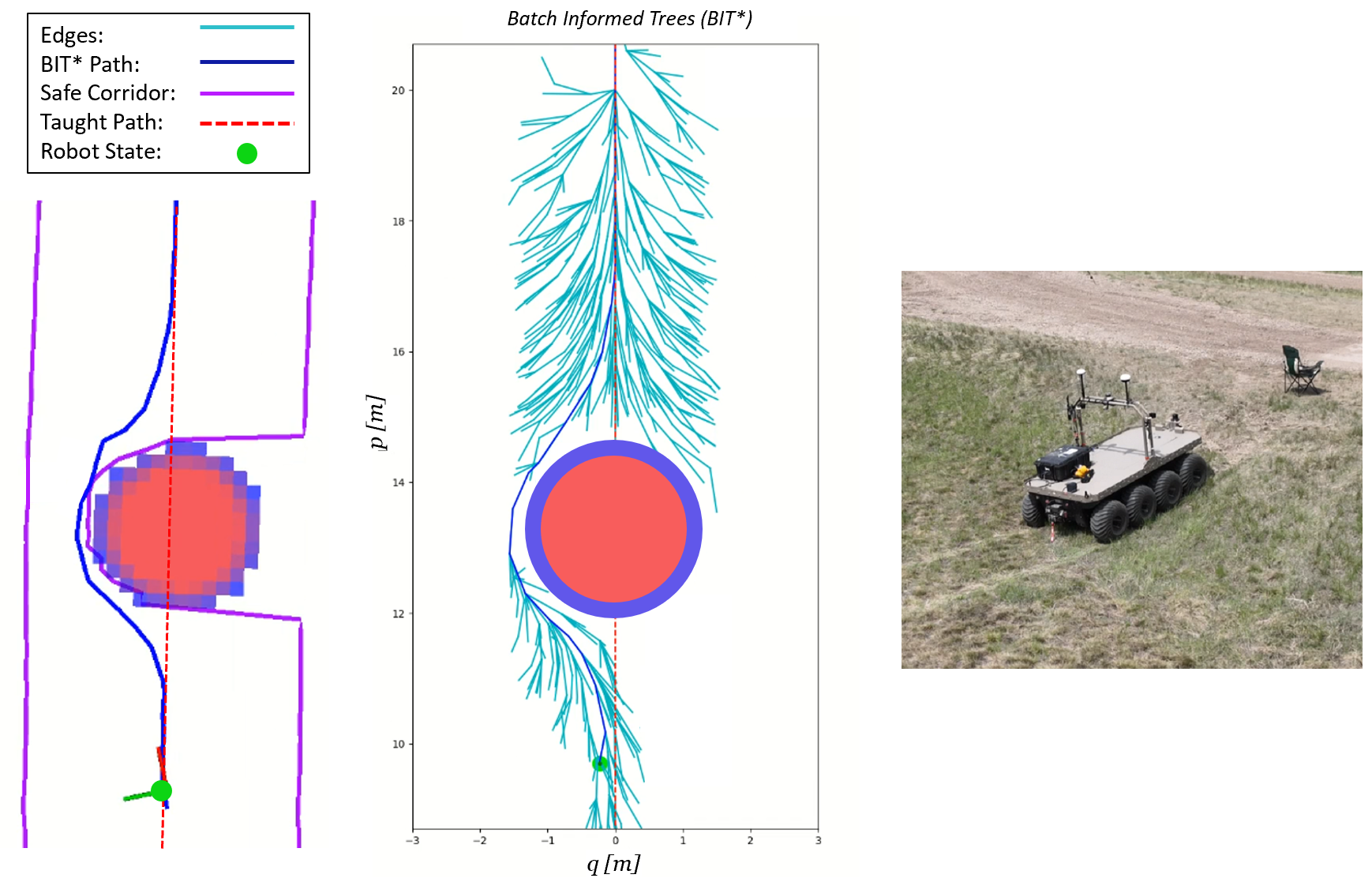}
	\caption[Hard Obstacle Avoidance Scenario with RViz Overlay]{A representative BIT* path and safe lateral corridor constraint solution, taken at a snapshot along the repeat in the hard scenario. As large swathes of the optimal solution lay on the reference path, it is often more practical to plan from the end of path to the robot state, as is shown here.}
	\label{OATS_Obs_Avoidance_Overlay}
\end{figure*}

In this experiment, a similar set of obstacles as in the easier-terrain scenario were used, but with more challenging placement, necessitating complex trajectories within a more heavily constrained safe corridor. Additionally, the presence of many elevation changes, shown in Fig. \ref{OATS_Obs_Repeat_Path}, obstructed the view of some obstacles along the repeat path until the vehicle approached close enough to clear occlusions, requiring faster replanning to successfully avoid obstacles. The total path length was 550 m, repeated 5 times for each motion planner, resulting in the robot encountering a total of 55 obstacles interactions over a cumulative navigation distance of 2.75 km. Fig. \ref{OATS_Obs_Avoidance_Overlay} shows a representative obstacle interaction and associated planner solution on the hard path with a chair placed at the crest of a steep slope.

In this more difficult test case, both motion planners successfully completed the repeats without explicit operator intervention. However, the direct-tracking \ac{MPC} exhibited minor glancing collisions with some obstacles due to kinematic tracking errors, resulting in an obstacle-avoidance rate of 87.27\%. In contrast, the homotopy-guided \ac{MPC} maintained a perfect 100\% obstacle-avoidance rate.

While the more successful obstacle-avoidance rates achieved by the homotopy-guided \ac{MPC} provide telling conclusions, the \ac{RMS} lateral and heading errors for the obstacle interactions further corroborate the earlier observations made during the easier obstacle course. The baseline obstacle-free path-tracking errors were measured as 6.12 cm for lateral errors and 6.1 deg for heading errors in the case of the direct-tracking \ac{MPC}. Similarly, the homotopy-guided \ac{MPC} exhibited obstacle-free lateral and heading errors of 5.94 cm and 6.0 deg, respectively. It is worth noting that the tracking errors increased in the baseline obstacle-free repeats compared to those of the easier environment, likely due to the more challenging reference path trajectories and inconsistent terrain composition. Following obstacle-free repeats, obstacles were introduced on the path and Table \ref{B15_Hard_Obs_Interaction_Table} reports the obstacle-interaction \ac{RMS} error metrics for the repeats.

\begin{table}[b]
	\vspace{-8pt}
	\caption{Obstacle Avoidance Error Table: Hard Scenario}
	\label{B15_Hard_Obs_Interaction_Table}
	\begin{center}
		\scalebox{0.75}{
			\begin{tabular}{|c|c||c|c||c|c|}
				\hline
				\multicolumn{2}{|c||}{} & \multicolumn{2}{c||}{\textbf{Direct-Tracking MPC}} & \multicolumn{2}{c|}{\textbf{Homotopy-Guided MPC}} \\
				\cline{3-6}
				\multicolumn{2}{|c||}{\specialcell[c]{\textbf{Obstacle \#}\\\textbf{Extent [m]}\\}} &
				\specialcell[c]{\textbf{Lateral}\\\textbf{RMSE [m]}} & \specialcell[c]{\textbf{Heading}\\\textbf{RMSE [deg]}} &
				\specialcell[c]{\textbf{Lateral}\\\textbf{RMSE [m]}} & \specialcell[c]{\textbf{Heading}\\\textbf{RMSE [deg]}} \\
				\hline
				1. & 0.1 & 0.31 & 8.48 & 0.22 & 4.32\\
				\hline
				2. & 0.1 & 0.38 & 13.49 & 0.33 & 10.20\\
				\hline
				3. & 0.5 & 0.70 & 21.81 & 0.49 & 13.39\\
				\hline
				4. & 0.1 & 0.40 & 17.40 & 0.28 & 10.18\\
				\hline
				5. & 0.3 & 0.52 & 13.45 & 0.37 & 11.70\\
				\hline
				6. & 0.3 & 0.40 & 15.39 & 0.36 & 11.80\\
				\hline
				7. & 0.6 & 0.58 & 17.61 & 0.54 & 11.10\\
				\hline
				8. & 0.7 & 0.69 & 19.40 & 0.63 & 13.01\\
				\hline
				9. & 0.4 & 0.50 & 14.88 & 0.35 & 10.56\\
				\hline
				10. & 1.0 & 0.81 & 17.82 & 0.83 & 14.84\\
				\hline
				11. & 0.8 & 0.58 & 12.37 & 0.60 & 9.76\\
				\hline
		\end{tabular}}
	\end{center}
	\vspace*{-0.35cm}
\end{table}

In addition to achieving more reliable and safe navigation, we consistently observe that the homotopy-guided \ac{MPC} maintains or surpasses the quality of local obstacle-avoidance trajectories compared to the direct-tracking \ac{MPC}, particularly in minimizing lateral and rotational path errors. On average, the direct-tracking \ac{MPC} deviates laterally from the reference path by a maximum of $r$ + 0.484 m, with a standard deviation of 0.158 m, while the homotopy-guided \ac{MPC} results in $r$ + 0.309 m, with a standard deviation of 0.052 m. It is noteworthy that the trajectory characteristics of the repeated obstacle-avoidance paths maintain the trends observed throughout the easy scenario, where the homotopy-guided \ac{MPC} exhibits slight advantages in terms of the consistency of lateral deviations, despite the significantly different operating environment. Both motion-planning solutions, however, generate qualitatively desirable paths that avoid obstacles while closely adhering to the reference path to complete the objective and exhibit consistent characteristics across a range of operating conditions.

When considering the average path curvature, we find further evidence supporting the superiority of the homotopy-guided \ac{MPC} over the direct-tracking approach. With an obstacle-free average path curvature of 0.106 m$^{-1}$, the direct-tracking method results in a robot trajectory with a curvature of 0.123 m$^{-1}$, while the homotopy-guided method achieves 0.114 m$^{-1}$. This corresponds to a change of 16.0\% and 7.5\%, respectively.

\section{Discussion and Conclusions}
In this work, we presented a local obstacle-avoidance architecture designed for path-following applications such as teach and repeat. By modifying a sample-based motion planner to use a laterally weighted edge-cost metric combined with a curvilinear planning space, we show how natural path following can be achieved that exploits prior knowledge of the terrain to avoid obstacles. Our method emphasizes reducing lateral deviations from a previously taught reference path, which is expected to offer increased safety during traversal. Critically, we demonstrate that through a novel preprocessing step, we can condition our path planner to elegantly handle infrequently occurring but troublesome rotation singularities in the reference paths without invalidating the underlying planner properties.

We then explored two \ac{MPC} architectures that leverage the planner solution to generate smooth robot trajectories and avoid local obstacles along the reference path. In the first method we elected to directly track the output path of the planner to produce a kinematically feasible robot trajectory. While simple and reasonably effective for simple problems, we find that due to kinematic tracking errors it is possible for collisions to occur unless this error is explicitly accounted for through obstacle inflation. By decoupling the \ac{MPC} from the planner using the proposed homotopy-guided dynamic lateral corridor state constraints, we show how we are able to find better obstacle-avoidance trajectories with significantly faster path solution times and a stronger guarantee on collision avoidance. 

Since our initial manuscript \cite{Sehn2022}, our motion planner has proven its versatility by being successfully deployed on various other autonomous platforms, ranging from indoor navigation robots \cite{Hugues2022}, to autonomous boats \cite{Huang2023}, with minimal tuning required between applications.

Our architecture is designed primarily for path following and teach-and-repeat applications, but it can also function as a traditional local motion planner. In this case, an intermittently updated global plan from a secondary planner could serve as the reference for generating the curvilinear space needed by our planner. A caveat is that any updates to the global plan would require re-initialization of the configuration space. However, this is generally not a costly process if done intermittently. This approach also addresses a potential limitation of our planner: if no solution can be found within the user-defined lateral bounds from the global plan, instead of stopping or requiring operator input, the global planner can be queried to update the plan, allowing the system to search for a new local trajectory in the adjusted configuration space.

We limit our planning configuration space to a maximum lateral boundary from the reference global path to ensure that solutions best exploit prior terrain knowledge and do not exceed the localization capabilities of \ac{VTR}. Inherently, this does prevent potential distant collision-free paths from being discovered, but we argue that these solutions may be too risky to pursue and opt to return no solution in these circumstances so that a manual operator can be flagged for temporary intervention. In alternative navigation stacks where localization is less problematic and there is higher risk tolerance, the lateral corridor can be either expanded or coupled with a global replanning phase to rectify this issue.

Future work might seek to validate the motion planner with dynamic obstacles and seek to make improvements to the architecture to shift to using predictive modeling of obstacles. While the proposed path planner does demonstrate responsiveness to dynamic obstacles, its reactive nature can lead to non-ideal trajectories as obstacles move and the optimal solution changes. To	address this, incorporating temporal planning and control techniques can be beneficial. By anticipating the movement of dynamic obstacles, the planner can reduce the number of replanning cycles and collaborate more effectively with other dynamic agents.

Fortunately, the proposed-motion planning architecture is well suited for this adaptation in future work. BIT* has been demonstrated to perform admirably in higher-dimensional planning spaces, and it is likely that adding a temporal dimension to the curvilinear configuration space would scale well to real-time operation. Our collision-checking scheme is also malleable to temporal collision checks. Currently, when we collision-check our planning edges, we convert the curvilinear points to Euclidean space and query a 2D OGM. In a curvilinear space with a temporal dimension, the approach is the same. However, in this case we propose the use of a so-called Spatiotemporal Occupancy Grid Map (SOGM) \cite{Hugues2022} that can be queried in both time and space for collision checking. With a perception module designed to estimate the SOGM’s and a temporal axis in the planning domain, no other architectural or algorithmic changes would be required to accommodate this upgrade.

\newcommand{\BIBdecl}{\setlength{\itemsep}{0.05 em}}
\bibliographystyle{IEEEtran}
\bibliography{IEEEabrv,bibliography_file}

\newpage
\onecolumn

\appendix

\renewcommand{\thesection}{\Alph{section}}
\section{Gauss Newton Solution Method For Constrained Optimization Problems Using Lie Groups}
From Section V, we wish to find a solution to the optimization problem defined by
\begin{subequations}
	\begin{align} \label{eq:starta}
	&\begin{aligned} [t]
	J(\T ,\ub, \y) =
	\sum_{k=1}^K \ln (\Trefk \Tk^{-1})^{\vee^T}
	\Matrix{Q}_k
	\ln (\Trefk \Tk^{-1})^{\vee}
	+ \ub_k^T \Matrix{R}_k \ub_k
	&+ \vel^T
	\, \V \, \vel\\
	&+ \lat^T \, \W \, \lat \\
	\end{aligned} \\
	&\begin{aligned} [b]
	&\mathrm{s.t.} \\
	&\Tkone =
	\exp \Big( (\Matrix{P}^{T} \uk)^\wedge h \Big) \Tk, \quad k = 1, 2, \hdots , K. \label{eq:startb}
	\end{aligned}
	\end{align}
\end{subequations}
Problem \eqref{eq:starta} can be solved efficiently with a Gauss-Newton method by linearizing the problem at an operating point and applying small perturbations to the Lie Algebra to simplify terms. In practice, we solve the \ac{MPC} problem directly using the \ac{STEAM} engine \cite{Anderson2015}, an iterative Gauss-Newton-style optimization library aimed at solving batch nonlinear optimization problems involving both Lie Group and Continuous-time components. However, as an exercise we show how one might attempt to manually solve this problem leveraging methods and concepts discussed in Barfoot (2017). 

We start by trying to linearize the summation terms about an operating point and perform Gauss Newton. Using a small perturbation in the Lie Algebra, $\ek$, we can represent the state $\Tk$ as
\begin{subequations}
	\begin{align}
	\Tk &= \dTk \Tkop \label{eq:twelve}\\
	\dTk &= \exp (\ek^\wedge). \label{eq:thirteen}
	\end{align}
\end{subequations}
\noindent For small $\ek$, we can then use Equations \eqref{eq:twelve} and \eqref{eq:thirteen} to make the approximation that
\begin{equation}\label{eq:fourteen}
\begin{aligned} [t]
\ln (\Trefk \Tk^{-1})^\vee &= \ln (\Trefk\Tkop^{-1} \exp (-\ek^\wedge))\\
& \approx \ekop + \pmb{\mathcal{J}}_l(-\ekop)^{-1}(-\ek).
\end{aligned}
\end{equation}
\noindent In this context, $\pmb{\mathcal{J}}_l$ is the left Jacobian matrix evaluated at $\ekop$, with
\begin{subequations}
	\begin{align}
	\pmb{\mathcal{J}}_l (\ekop) &= \sum_{n=0}^{\infty}
	\frac{1}{(n+1)!} (\ekop^\curlywedge)^n \label{eq:fifteen}\\
	\ekop &= \ln (\Trefk \Tkop^{-1})^\vee. \label{eq:sixteen}
	\end{align}
\end{subequations}
We then make the following definitions to raise the first term of \eqref{eq:starta} into lifted form:
\begin{subequations}
	\begin{align}
	&\ebarop = \bbm \Matrix{\epsilon}_{1,\mathrm{op}} \\
	\Matrix{\epsilon}_{2,\mathrm{op}} \\ \vdots \\ \Matrix{\epsilon}_{K,\mathrm{op}} \ebm, \qquad
	\ebar = \bbm \Matrix{\epsilon}_1 \\
	\Matrix{\epsilon}_2 \\ \vdots \\
	\Matrix{\epsilon}_K \ebm, \label{eq:seventeena}\\
	&\F =  \mathrm{diag}(\pmb{\mathcal{J}}_r(\Matrix{\epsilon}_{1,\mathrm{op}})^{-1}, \pmb{\mathcal{J}}_r(\Matrix{\epsilon}_{2,\mathrm{op}})^{-1}, \hdots , \pmb{\mathcal{J}}_r(\Matrix{\epsilon}_{K,\mathrm{op}})^{-1}), \label{eq:seventeeca}\\
	& \Q = \mathrm{diag}(\Q_1, \Q_2, \hdots , \Q_K). \label{eq:seventeend}
	\end{align}
\end{subequations}
This results in the representation of the summation term in \eqref{eq:starta} as
\begin{align}
\begin{aligned}\label{eq:eighteen}
\sum_{k=1}^K \ln (\Trefk \Tk^{-1})^{\vee^T} &\Q_k
\ln (\Trefk \Tk^{-1})^{\vee} = (\ebarop - \F \ebar)^T
\Q (\ebarop - \Matrix{F} \ebar).
\end{aligned}
\end{align}
For the penalty terms $\vel$ and $\lat$, we use a two-term Taylor expansion, letting $\ub = \uop + \du$ with
\begin{subequations}
	\begin{align}
	&\du = \bbm \du_1 \\ \du_2 \\ \vdots \\ \du_K \ebm, \qquad
	\uop = \bbm \ub_{1,\mathrm{op}} \\ \ub_{1,\mathrm{op}} \\ \vdots \\ \ub_{K,\mathrm{op}} \ebm, \label{eq:nineteenb} \\
	&\begin{aligned} [b] \label{eq:nineteenc}
	\Matrix{\Phi}_{\mathrm{vel}}(\uop + \du) &\approx \velop +
	\Big( \frac{\partial \, \vel}{\partial 	\Matrix{u} } \bigg\rvert_{\uop} \Big) \du \\
	&\approx \velop + \M \du.
	\end{aligned}
	\end{align}
\end{subequations}
We define $\M$ as the Jacobian of $\vel$ evaluated at $\uop$. For the case of $\lat$, we let
\begin{align} \label{eq:twenty}
\yk &= \IdentityMatrix_{2}^{T} \Trefk \Tk^{-1} \IdentityMatrix_{4}
= \ykop + \dyk,
\end{align}
where $\IdentityMatrix_{i}$ is the $ith$ column of the identity matrix. Once again we apply the perturbation scheme of \eqref{eq:twelve} and using the approximation that $\ek$ is small results in the simplification that
\begin{equation} \label{eq:twentyone}
\begin{aligned} [t]
\Trefk \Tk^{-1} &= \Trefk \Tkop^{-1} \dTk^{-1} \\
& = \Trefk \Tkop^{-1} \exp(-\ek^\wedge) \\
& \approx \Trefk \Tkop^{-1} (\IdentityMatrix -\ek^\wedge).
\end{aligned}
\end{equation}
Substituting \eqref{eq:twentyone} into \eqref{eq:twenty} and distributing terms allows us to represent $\ykop$ and $\dyk$ as
\begin{align}
&\ykop = \IdentityMatrix_{2}^{T} \Trefk \Tkop^{-1} \IdentityMatrix_{4} \label{eq:twentytwo}
\end{align}
and
\begin{align}
&\dyk =  -\IdentityMatrix_{2}^{T} \Trefk \Tkop^{-1} \ek^\wedge \IdentityMatrix_{4}, \label{eq:twentythree}
\end{align}
respectively. To simplify the skew-symmetric operator on $\ek$, we make use of the overloaded operator $\odot$ and an identity defined by the equations
\begin{subequations}
	\begin{align}
	&\Matrix{\epsilon}^\wedge \Matrix{p} = \Matrix{p}^\odot \Matrix{\epsilon} \label{eq:twentyfoura}\\
	&\Matrix{p}^\odot = \bbm \Matrix{\rho} \\ \eta \ebm^{\odot} =
	\bbm \eta \IdentityMatrix & -\Matrix{\rho}^\wedge \\ \ZeroMatrix^T & \ZeroMatrix^T \ebm. \label{eq:twentyfourb}
	\end{align}
\end{subequations}
Applying \eqref{eq:twentyfoura} to \eqref{eq:twentythree} yields
\begin{align}
&\Matrix{p} =  \IdentityMatrix_{4} \label{eq:twentyfive}\\
& \dyk = -\IdentityMatrix_{2}^{T} \Trefk \Tkop^{-1} \Matrix{p}^\odot \ek = \Matrix{g}_k^T \ek, \label{eq:twentysix}
\end{align}
where,
\begin{align}
&\Matrix{g}_k^T = -\IdentityMatrix_{2}^{T} \Trefk \Tkop^{-1} \Matrix{p}^\odot. \label{eq:twentyseven}
\end{align}
Then, we can raise $\ykop$ and $\dyk$ to lifted form using new definitions of $\G$, $\yop$, and $\dy$ given by
\begin{subequations}
	\begin{align}
	&\G = \bbm \Matrix{g}_1^T & & & \\ 
	& \Matrix{g}_2^T & & \\
	& & \ddots & \\
	& & & \Matrix{g}_K^T \ebm, \qquad
	\yop = \bbm \y_{1,\mathrm{op}} \\ \y_{2,\mathrm{op}} \\ \vdots \\ \y_{K,\mathrm{op}} \ebm, \qquad 
	\dy = \bbm \dy_{1} \\ \dy_{2} \\ \vdots \\ \dy_{K} \ebm = \G \ebar. \label{eq:twentyeightc}
	\end{align}
\end{subequations}
Returning to $\lat$, we now have a two-term Taylor expansion,
\begin{equation} \label{eq:twentynine}
\begin{aligned} [t]
\Matrix{\Phi}_{\mathrm{lat}}(\yop + \dy) &\approx \latop +
\Big( \frac{\partial \, \lat}{\partial \y } \bigg\rvert_{\yop} \Big) \dy \\
&\approx \latop + \N \G \ebar
\end{aligned}
\end{equation}
that can be used to simplify the lateral barrier constraint. In this context, $\N$ is the Jacobian of $\lat$ evaluated at $\yop$. Now we can reassemble a quadratic approximation of the original cost function \eqref{eq:starta} by substituting in \eqref{eq:eighteen}, \eqref{eq:nineteenc}, and \eqref{eq:twentynine}, yielding
\begin{subequations}
	\begin{align} \label{eq:thirtya}
	&\begin{aligned} [t]
	J(\ebar,\du) &=
	(\ebarop - \F \ebar)^T
	\Q (\ebarop - \F \ebar) + (\uop +\du)^T \R (\uop +\du)\\
	&+ \Big( \velop + \M \du \Big)^T
	\V \Big( \velop + \M \du \Big)\\
	&+ \Big(\latop + \N \G \ebar \Big)^T \W \Big(\latop + \N \G \ebar \Big)
	\end{aligned} \\
	&\begin{aligned} [b]
	&\mathrm{s.t.} \\
	&\Tkone = \exp \Big( \Puk^\wedge h \Big) \Tk, \quad k = 1, 2, \hdots , K, \label{eq:thirtyb}
	\end{aligned}
	\end{align}
\end{subequations}
where $\Matrix{Q}$, $\Matrix{R}$, $\Matrix{V}$, and $\Matrix{W}$ are tunable diagonal weighting matrices. We note that \eqref{eq:thirtya} is approximated as quadratic in terms of two optimization variables. Next we work with the equality constraint \eqref{eq:thirtyb} to try to find a relationship between $\ebar$ and $\dubar$. Once more leveraging the perturbation scheme of \eqref{eq:twelve}, \eqref{eq:thirteen} and substituting $\uk = \ukop + \duk$ we find that
\begin{equation}
\begin{aligned} [b]
\Tkone &= \exp \Big( \Puk^{\wedge} h \Big) \Tk \\
\exp \Big(\ekone^\wedge \Big) \Tkoneop &= 
\exp \Big( \big(\Pukop + \Pudk \big)^\wedge h \Big) \exp(\ek^\wedge)\Tkop.
\label{eq:thirtyone}
\end{aligned}
\end{equation}
Next we leverage the approximation and identity \cite{Barfoot2017}
\begin{subequations}
	\begin{align}
	&\exp \Big((\Matrix{x} + \delta \Matrix{x})^\wedge \Big) \approx
	\exp \Big((\pmb{\mathcal{J}}_l (\Matrix{x}) \delta \Matrix{x})^\wedge \Big)
	\exp(\Matrix{x}^\wedge) \label{eq:thirtytwo}\\
	&\T \exp(\Matrix{x}^\wedge) = \exp \Big( (\pmb{\mathcal{T}} \Matrix{x} \ )^\wedge \Big) \T.
	\label{eq:thirtythree}
	\end{align}
\end{subequations}
to try to simplify \eqref{eq:thirtyone}. First we use \eqref{eq:thirtytwo} to expand the large exponential term of \eqref{eq:thirtyone} to be
\begin{align}
\exp \Big(\ekone^\wedge \Big) \Tkoneop = 
\exp \Big( \big(\pmb{\mathcal{J}}_l\Pukop \Pudk \big)^\wedge h \Big)
\exp \Big( \Pukop^\wedge h \Big)	
\exp(\ek^\wedge)\Tkop.
\label{eq:thirtyfour}
\end{align}
Now we can flip the order of the $\exp \Big( \Pukop^\wedge h \Big)$ and $\exp (\ek^\wedge)$ terms using \eqref{eq:thirtythree} resulting in the equivalent expression
\begin{equation}
\begin{aligned} [b]
&\exp \Big(\ekone^\wedge \Big) \Tkoneop \\
&= \exp \Big( \big(\pmb{\mathcal{J}}_l\Pukop \Pudk \big)^\wedge h \Big)
\exp \Big( \Big( \exp \big( \Pukop^\curlywedge h \big) \ek \Big)^\wedge \Big)
\exp \Big( \Pukop^\wedge h \Big) \Tkop. \label{eq:thirtyfive}
\end{aligned}
\end{equation}
From \eqref{eq:thirtyb} we see that the right hand terms of each side of equation \eqref{eq:thirtyfive} will cancel. Furthermore, we introduce the approximation that for small perturbations $\ek$, $\exp (\ek^\wedge) \approx \IdentityMatrix + \ek^\wedge$ we can simplify 
\begin{align} \label{eq:thirtysix}
&\IdentityMatrix + \ekone^\wedge
= \Big(  \IdentityMatrix + \big(\pmb{\mathcal{J}}_l\Pukop \Pudk \big)^\wedge h \Big)
\Big( \IdentityMatrix + \Big( \exp \big( \Pukop^\curlywedge h \big) \ek \Big)^\wedge \Big),
\end{align}
where $\IdentityMatrix$ is the identity matrix.
Noting that both $\ek$ and $\duk$ are small, we distribute the terms of the polynomial and approximate the cross term as zero. The identity matrices cancel on both sides, and the inverse of the skew-symmetric operator is applied to both sides resulting in
\begin{align} \label{eq:thirtyseven}
&\ekone
\approx \pmb{\mathcal{J}}_l \Pukop \Pudk h +
\exp \big( \Pukop^\curlywedge h \big) \ek.
\end{align}
If we then define $\A_k$ and $\B_k$ as
\begin{subequations}
	\begin{align}
	&\A_k = \pmb{\mathcal{J}}_l \Pukop h \label{eq:thirtyeighta} \\
	&\B_k = \exp \big( \Pukop^\curlywedge h \big) \label{eq:thirtyeightb}
	\end{align}
\end{subequations}
we have the simplification that
\begin{align} \label{eq:thirtynine}
\Matrix{P}^{T} \duk = -\A_k^{-1} \B_k \ek + \A_k^{-1} \ekone.
\end{align}
If we define the matrix $\Hb$ as
\begin{align} \label{eq:forty}
\Hb =  \bbm -\A_1^{-1} \B_1 & \A_1^{-1} & &  \\
& \ddots & \ddots &  \\
& & -\A_{K-1}^{-1} \B_{K-1} & \A_{K-1}^{-1} \\
& & & -\A_{K}^{-1} \B_{K} 
\ebm,
\end{align}
we can compactly represent \eqref{eq:thirtynine} for all $k$ as 
\begin{subequations}
	\begin{align}
	\Matrix{P}^T \du &= \Hb \ebar, \label{eq:fortyonea} \\
	\Hb^{-1} \Matrix{P}^T \du &= \ebar. \label{eq:fortyoneb}
	\end{align}
\end{subequations}
Lastly, we use the result of \eqref{eq:fortyoneb}, to relate $\du$ and $\ebar$ such that we can re-write the quadratic approximation of our cost function \eqref{eq:thirtya} in terms of a single optimization variable resulting in
\begin{equation} \label{eq:fortytwo}
\begin{aligned} [t]
J(\du) &=
(\ebarop - \F \Hb^{-1} \Matrix{P}^T \du)^T
\Q (\ebarop - \F \Hb^{-1} \Matrix{P}^T \du)
+ (\uop + \du)^T \R (\uop + \du)\\
&+ \Big( \velop + \M \du \Big)^T
\V \Big( \velop + \M \du \Big)\\
&+ \Big(\latop + \N \G \Hb^{-1} \Matrix{P}^T \du \Big)^T \W \Big(\latop + \N \G \Hb^{-1} \Matrix{P}^T \du \Big).
\end{aligned}
\end{equation}
Note that given an operating point, the function is quadratic in $\du$ and all other quantities are known. We minimize the function by taking the derivative and equating to zero, then solve the linear system of equations for $\du^*$. Finally, we update the operating point with
\begin{align} \label{eq:fortythree}
\uop \leftarrow \uop + \du^*
\end{align} 
and then continue this Gauss Newton iteration process until convergence of $\du^*$ below a threshold parameter and return the solution.

\end{document}